\documentclass[10pt,journal,compsoc]{IEEEtran}
\pdfoutput=1

\usepackage{times}
\usepackage{graphicx}
\usepackage{amsmath}
\usepackage{amssymb}
\usepackage{enumitem}
\usepackage{subcaption} 
\usepackage{varwidth} 
\usepackage{breqn}
\usepackage{epstopdf}
\usepackage{tabulary} 
\usepackage{multirow} 
\usepackage{hhline} 
\usepackage{amsthm}  
\usepackage{cite}

\newtheorem{theorem}{Theorem}

\usepackage{url}            
\usepackage{booktabs}       
\usepackage{amsfonts}       
\usepackage{placeins} 
\usepackage{nicefrac}       
\usepackage{microtype}      

\usepackage{xcolor}
\usepackage{tikz}
\usetikzlibrary{arrows,shapes,positioning}

\newcommand{\noarxiv}[1]{}

\usepackage{accents}
\newlength{\dhatheight}

\newcommand{\blankline}{\vspace{\baselineskip}} 

\def\mD{\mathcal{D}}
\def\mE{\mathcal{E}}

\def\mI{\mathcal{I}}

\def\mL{\mathcal{L}}
\def\mM{\mathcal{M}}

\def\mP{\mathcal{P}}

\def\1n{\mathbf{1}_n}
\def\0{\mathbf{0}}
\def\1{\mathbf{1}}

\def\R{{\mathbb R}}

\def\n{{\bf n}}

\def\u{{\bf u}}

\def\x{{\bf x}}
\def\y{{\bf y}}
\def\z{{\bf z}}

\def\btheta{\mbox{\boldmath{$\theta$}}}

\def\tt{\mbox{\tiny $T$}}

\newcommand{\cX}{\mathcal{X}}
\newcommand{\cM}{\mathcal{M}}
\newcommand{\mbf}[1]{\mathbf{#1}}

\newcommand{\bv}{\mathbf{v}}
\newcommand{\cP}{\mathcal{P}}
\newcommand{\bu}{\mathbf{u}}

\DeclareMathOperator*{\argmin}{arg\,min}

\newcommand{\sigmoid}{\sigma}

\newcommand{\prox}{\text{\textbf{prox}}}

\newcommand{\iterk}[1]{ {#1}^{(k)} }
\newcommand{\iterkplus}[1]{ {#1}^{(k{+}1)} }

\usepackage{physics}

\usepackage{xspace}
\makeatletter
\DeclareRobustCommand\onedot{\futurelet\@let@token\@onedot}
\def\@onedot{\ifx\@let@token.\else.\null\fi\xspace}

\def\eg{\emph{e.g}\onedot} 
\def\ie{\emph{i.e}\onedot} 
 
\def\etc{\emph{etc}\onedot} \def\vs{\emph{vs}\onedot}
\def\wrt{w.r.t\onedot} 
\def\etal{\emph{et al}\onedot}
\makeatother

\usepackage[pagebackref=true,breaklinks=true,letterpaper=true,colorlinks,bookmarks=false]{hyperref}

\def\fiveimgwidth{0.19\linewidth}

\def\twelveimgwidth{0.085\linewidth}

\def\twelveimgrowmargin{-2mm}

\def\thirteenimgwidth{0.07\linewidth}

\def\thirteenimgrowcaptionheight{-8.5mm}
\def\thirteenimgrowmargin{-2mm}

\def\fifteenimgwidth{0.06\linewidth}

\def\fifteenimgrowcaptionheight{-10mm}
\def\fifteenimgrowmargin{-2mm}

\def\sixteenimgwidth{0.057\linewidth}

\def\sixteenimgrowcaptionheight{-9.5mm}
\def\sixteenimgrowmargin{-2mm}

\def\groundtruthrowcaption{\centering \scriptsize ground truth/  input}
\def\oursrowcaption{\centering \scriptsize  proposed}
\def\lrowcaption{\centering \scriptsize  $\ell_1$ prior}
\def\srowcaption{\centering \scriptsize  specially-trained network}

\def\Cblankcol{\centering \scriptsize $\qquad$}   
\def\Ccscol{\centering \scriptsize compressive sensing}   
\def\Cdenoisecol{\centering \scriptsize pixelwise inpaint, denoise}  
\def\Cblockcol{\centering \scriptsize blockwise inpaint}     
\def\Cinpaintcol{\centering \scriptsize scattered inpaint}  
\def\Csuperrescola{\centering \scriptsize $2\times$super-resolution}  
\def\Csuperrescolb{\centering \scriptsize $4\times$super-resolution}  

\newcommand{\myparagraph}[1]{ \blankline \noindent \textbf{#1} \hspace{2mm}}

\begin{document}

	\title{\LARGE \textit{One Network to Solve Them  All} --- Solving  Linear Inverse Problems using Deep Projection Models}
	

	\author{J.\ H.\ Rick Chang\thanks{Chang, Bhagavatula and Sankaranarayanan were supported, in part, by the ARO Grant W911NF-15-1-0126. Chang was also partially supported by the CIT Bertucci Fellowship.}, Chun-Liang Li, Barnab{\'a}s P{\'o}czos, B.\ V.\ K.\ Vijaya Kumar, \\ 
		and Aswin C.\ Sankaranarayanan \\
		Carnegie Mellon University, Pittsburgh, PA\\
	}

	\IEEEtitleabstractindextext{%
		
\begin{abstract}

While deep learning methods have achieved state-of-the-art performance in many challenging  inverse problems like image inpainting and super-resolution, they invariably involve problem-specific training of the networks. 
Under this approach, different problems require different networks.
In scenarios where we need to solve a wide variety of problems, \eg, on a mobile camera, it is inefficient and costly to use these specially-trained networks.
On the other hand, traditional methods using signal priors can be used in all linear inverse problems but often have worse performance on challenging tasks.
In this work, we provide a middle ground between the two kinds of methods --- we propose a general framework to train a single deep neural network that solves arbitrary linear inverse problems.
The proposed network acts as a proximal operator for an optimization algorithm and projects non-image signals onto the set of natural images defined by the decision boundary of a classifier.
In our experiments, the proposed framework demonstrates superior performance over traditional methods using a wavelet sparsity prior and achieves comparable performance of specially-trained networks on tasks including compressive sensing and pixel-wise inpainting. 

\end{abstract}

		\begin{IEEEkeywords}
			Linear inverse problems, Image processing, Adversarial learning, Proximal operator
		\end{IEEEkeywords}
	}

	\maketitle

\def\dataset{ring_512}
\def\denoise{denoise_ratio0.80_std0.00}
\def\denoiseours{ours_alpha0.150000}
\def\cs{cs_ratio0.10_std0.00}
\def\csours{ours_alpha0.300000}
\def\inpaint{inpaint_bs4_tb5000_std0.00}
\def\inpaintours{ours_alpha0.040000}
\def\superres{superres_ratio0.50_std0.00}
\def\superresours{ours_alpha0.500000}

\def\hshift{-10mm}

\begin{figure*}[t]
	\hspace{\hshift}
	\centering
	\begin{subfigure}[t]{0.1\linewidth}
		\caption*{\Cblankcol}
	\end{subfigure}	
	\begin{subfigure}[t]{\fiveimgwidth}
		\centering
		\caption*{\centering \scriptsize{ compressive sensing \hspace{\linewidth} ($10\times$ compression)}  }
	\end{subfigure}	
	\begin{subfigure}[t]{\fiveimgwidth}
		\centering
		\caption*{\centering \scriptsize{ pixelwise inpainting and denoising  ($80\%$ drops)}}
	\end{subfigure}	
	\begin{subfigure}[t]{\fiveimgwidth}
		\centering
		\caption*{\centering \scriptsize{ \newline \hspace{\linewidth} scattered inpainting }}
	\end{subfigure}	
	\begin{subfigure}[t]{\fiveimgwidth}
		\centering
		\caption*{\centering \scriptsize{ \newline \hspace{\linewidth} $2\times$  super-resolution }}
	\end{subfigure}	
	\\
	\hspace{\hshift}
	\begin{subfigure}[t]{0.1\linewidth}
		\centering
		\vspace{-15mm}
		\caption*{\centering \scriptsize{ground truth / input}}
	\end{subfigure}	
	\begin{subfigure}[t]{\fiveimgwidth}
		\includegraphics[width=\textwidth]{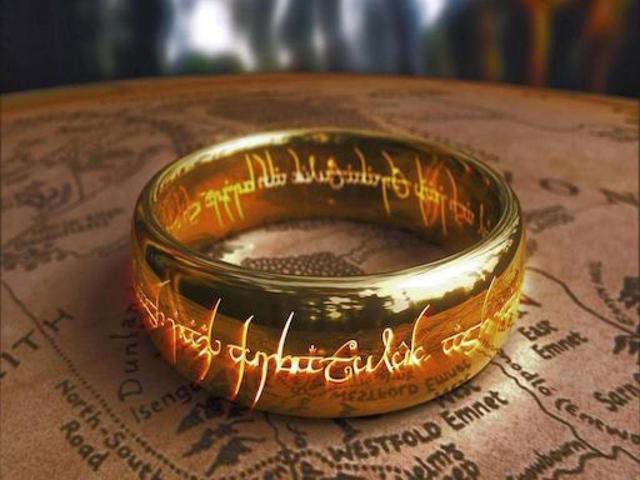}
	\end{subfigure}	
	\begin{subfigure}[t]{\fiveimgwidth}
		\includegraphics[width=\textwidth]{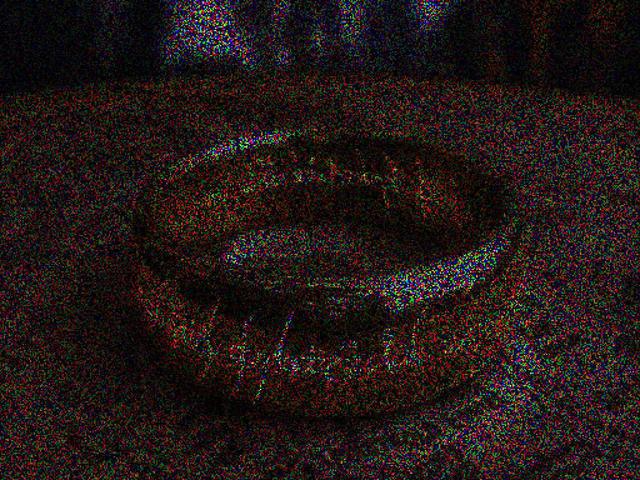}
	\end{subfigure}	
	\begin{subfigure}[t]{\fiveimgwidth}
		\includegraphics[width=\textwidth]{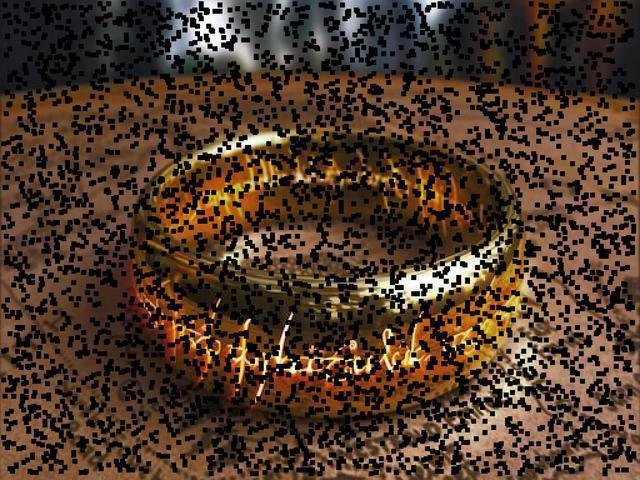}
	\end{subfigure}
	\begin{subfigure}[t]{\fiveimgwidth}
		\centering
		\vspace{-0.55\textwidth}
		\includegraphics[width=0.5\textwidth]{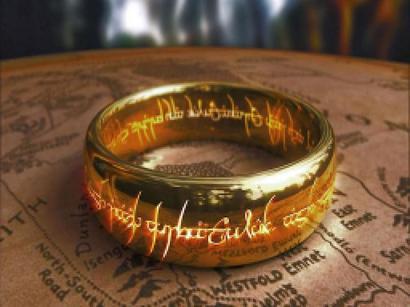}
	\end{subfigure}	
	\\
	\vspace{1mm}
	\hspace{\hshift}
	\begin{subfigure}[t]{0.1\linewidth}
		\centering
		\vspace{-15mm}
		\caption*{\centering \scriptsize{reconstruction output}}
	\end{subfigure}	
	\begin{subfigure}[t]{\fiveimgwidth}
		\includegraphics[width=\textwidth]{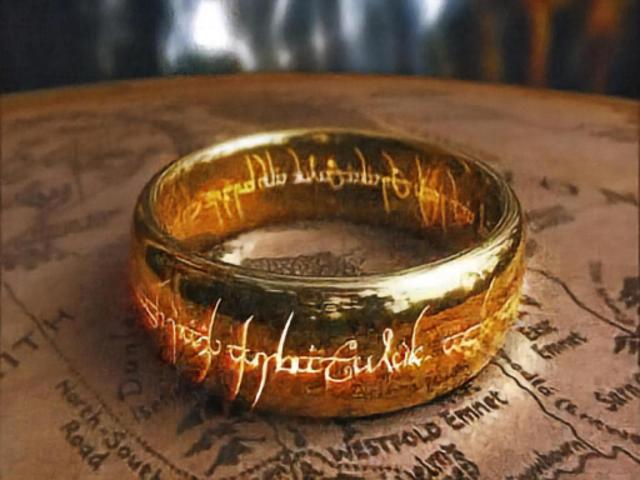}
	\end{subfigure}	
	\begin{subfigure}[t]{\fiveimgwidth}
		\includegraphics[width=\textwidth]{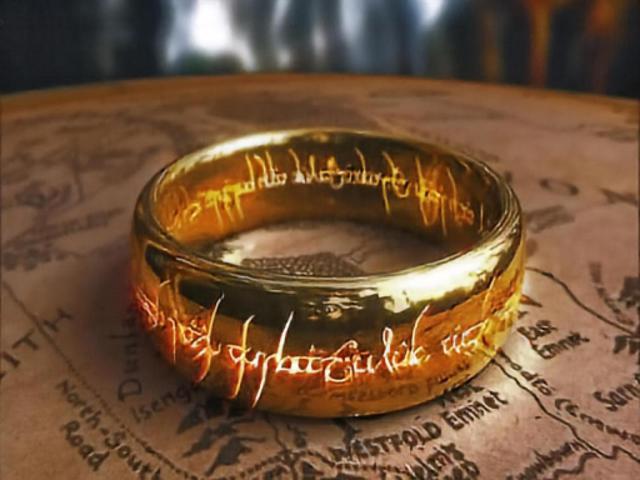}
	\end{subfigure}	
	\begin{subfigure}[t]{\fiveimgwidth}
		\includegraphics[width=\textwidth]{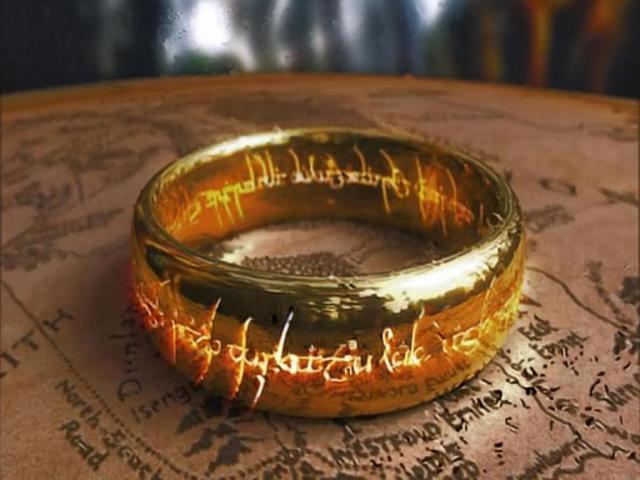}
	\end{subfigure}	
	\begin{subfigure}[t]{\fiveimgwidth}
		\includegraphics[width=\textwidth]{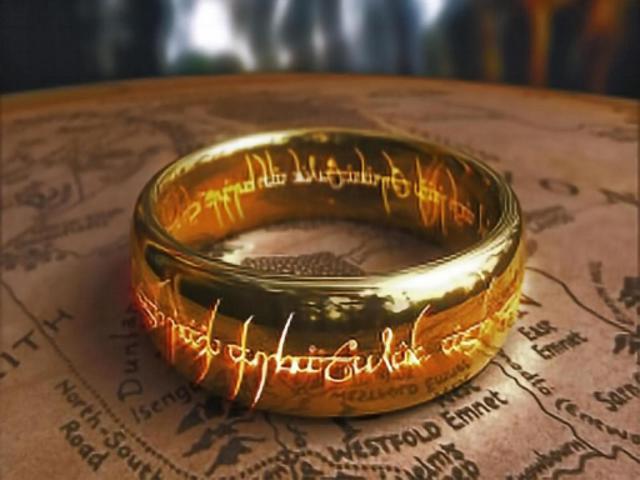}
	\end{subfigure}
	\caption{The same proposed network is used to solve the following tasks: compressive sensing problem with $10\times$ compression, pixelwise random inpainting with $80\%$ dropping rate, scattered inpainting, and $2\times$-super-resolution.  Note that even though the nature and input dimensions of the problems are very different, the proposed framework is able to use a single network to solve them all without retraining. }
	\label{figure: venice}
\end{figure*}

\section{Introduction}

At the heart of many image processing tasks is a linear inverse problem, where the goal is to reconstruct an image $\x \in \R^d$ from a set of measurements $\y \in \R^m$ of the form $\y = A \x + \n$, where $A\in\R^{m\times d}$ is the measurement operator and $\n \in \R^m$ is the noise.
For example, 
in image inpainting, $\y$ is an image with masked regions and $A$ is the linear operation applying a pixelwise mask to the original image $\x$; 
in super-resolution, $\y$ is a low-resolution image and the operation $A$ downsamples high resolution images; 
in compressive sensing, $\y$ denotes compressive measurements and $A$ is the measurement matrix, \eg, a random Gaussian matrix.  
%
%
Linear inverse problems, like those described above, are often underdetermined, \ie, they involve fewer measurements than unknowns. 
Such underdetermined systems are extremely difficult to solve since the operator $A$ has a non-trivial null space and there are an infinite number of feasilbe solutions but only a few of them are natural images.

%
%

\myparagraph{Solving linear inverse problems.} 
There are two broad classes of methods for solving linear underdetermined problems. 
At one end, we have techniques that use signal priors to regularize the inverse problems.
%
%
%
Signal priors enable identification of the true solution from the infinite set of feasible solutions by enforcing image-specific features to the solution.
%
%
%
%
%
%
%
%
%
Thereby, designing a signal prior plays a key role when solving linear inverse problems.
Traditionally, signal priors are hand-designed based on empirical observations of images. 
%
%
For example, since natural images are usually sparse after wavelet transformation and 
are generally piecewise smooth, 
signal priors that constrain the sparsity of wavelet coefficients or spatial gradients are widely used~\cite{donoho1995noising,portilla2003image,mairal2008learning,chan2006total,dong2011image}.
%
%
Even though these signal priors can be used in any linear inverse problems related to images and usually have efficient solvers, these signal priors are often too generic, in that many non-image signals can also satisfy the constraints.
Thereby, these hand-designed signal priors cannot easily deal with challenging problems like image inpainting or super-resolution.  
%
%

Instead of using a universal signal prior, a second class of method  learn a mapping from the linear measurement domain of $\y$ to the image space of $\x$, with the help of large datasets and deep neural nets~\cite{dong2014learning,ledig2016photo,pathak2016context}.
For example, to solve image super-resolution, low-resolution images are generated from a high-resolution image dataset, and the mapping between the corresponding image pairs are learned with a neural net~\cite{dong2014learning}.
Similarly, a network can be trained to solve compressive sensing problem~\cite{mousavi2017learning,kulkarni2016reconnet,mousavi2015deep} or image debluring~\cite{xu2014deep}, \etc.  
These methods have achieved state-of-the-art performance in many challenging problems.

%
%
%
Despite their superior performance, these specially-trained solvers are designed for specific problems and usually cannot solve other problems without retraining the mapping function --- even when the problems are similar. 
For example, a $2\times$-super-resolution network cannot be easily readapted to solve $4\times$ or $8\times$ super-resolution problems; a compressive sensing network for Gaussian random measurements is not  applicable to sub-sampled Hadamard measurements.  
%
%
%
%
Training a new network for every single instance of an inverse problem is a wasteful proposition.
In comparison, traditional methods using hand-designed signal priors can solve any linear inverse problems but have poorer performance on an individual problem.
Clearly, a middle ground between these two classes of methods is needed.

\myparagraph{One network to solve them all.}
In this paper, we ask the following question: \textit{if we have a large image dataset, can we learn from the dataset a signal prior that can deal with \textbf{any} linear inverse problems of images? }
Such a signal prior can significantly lower the cost to incorporate inverse algorithms into consumer products, for example, via the form of specialized hardware design.
%
%
To answer this question, we observe that in optimization algorithms for solving linear inverse problems, signal priors usually appears in the form of proximal operators. 
%
%
Geometrically, the proximal operator \emph{projects} the current estimate closer to the feasible sets (natural images) constrained by the signal prior.  
%
%
Thus, we propose to learn the proximal operator with a \emph{deep projection model}, which can be integrated into many standard optimization frameworks for solving arbitrary linear inverse problems.
%
%
%
%
%
%

\myparagraph{Contributions.}
We make the following contributions.

\begin{itemize}[leftmargin=*]
	\item We propose a general framework that implicitly learns a signal prior and a projection operator from large image datasets.  When integrated into an alternating direction method of multipliers (ADMM) algorithm, the same proposed projection operator can solve challenging linear inverse problems related to images.
	\item We identify sufficient conditions for the convergence of the nonconvex ADMM with the proposed projection operator, and we use these conditions as guidelines to design the proposed  projection network.
	\item We show that it is inefficient to solve generic linear inverse problems with state-of-the-art methods using specially-trained networks.   Our experiment results also show that they are prone to be affected by changes in the linear operators and noise in the linear measurements. In contrast, the proposed method is more robust to these factors.
\end{itemize}

	\section{Related Work}


Given noisy linear measurements $\y$ and the corresponding linear operator $A$, which is usually underdetermined, the goal of linear inverse problems is to find a solution $\x$, such that $\y \approx A \x$ and $\x$ be a signal of interest, in our case, an image.
Based on their strategies to deal with the underdetermined nature of the problem, algorithms for linear inverse problems can be roughly categorized into those using hand-designed signal priors and those learning from datasets.
We briefly review some of these methods.

\myparagraph{Hand-designed signal priors.}
%
Linear inverse problems are usually regularized by signal priors in a penalty form:
\begin{equation}
\min_{\x} \  \frac{1}{2} \norm{\y - A \x }_2^2 + \lambda \phi(\x), 
\label{eq: inverse}
\end{equation}
where $\phi: \R^d \rightarrow \R$ is the signal prior and $\lambda$ is the non-negative weighting term.
%
%
Signal priors constraining the sparsity of $\x$ in some transformation domain have been widely used in literatures. 
For example, since images are usually sparse after wavelet transformation or after taking gradient operations, a signal prior $\phi$ can be formulated as $\phi(\x) = \norm{W \x}_1$, where $W$ is a operator representing either wavelet transformation, taking image gradient, or other hand-designed linear operation that produces sparse features from images~\cite{donoho1998data}.
Using signal priors of $\ell_1$-norms enjoys two advantages.
First, it forms a convex optimization problem and provides global optimality.  
The optimization problem can be solved efficiently with a variety of algorithms for convex optimization. 
Second, $\ell_1$  priors enjoy many theoretical guarantees, thanks to results in compressive sensing~\cite{candes2006stable}.
For example, if the linear operator $A$ satisfies conditions like the restricted isometry property and $W \x$ is sufficiently sparse, the optimization problem~\eqref{eq: inverse} provides the sparsest solution.
%

%
Despite their algorithmic and theoretical benefits, hand-designed priors are often too generic to constrain the solution set of the inverse problem~\eqref{eq: inverse} to be images --- we can easily generate noise signals that have sparse wavelet coefficients or gradients. 
%

\myparagraph{Learning-based methods.}
The ever-growing number of images on the Internet enables state-of-the-art algorithms to deal with challenging problems that traditional methods are incapable of solving.
%
%
%
For example, image inpainting and restoration can be performed by pasting image patches or transforming statistics of pixel values of similar images in a large dataset~\cite{hays2007scene,dale2009image}.
Image denoising and super-resolution can be performed with dictionary learning methods that reconstruct image patches with sparse linear combinations of dictionary entries learned from datasets~\cite{aharon2006rm,yang2010image}.
%
%
%
%
Large datasets can also help learn end-to-end mappings from the linear measurement domain to the image domain.
%
Given a linear operator $A$ and a dataset $\cM = \{\x_1, {\dots}, \x_n\}$,  the pairs $\left\{(\x_i,
A\x_i)\right\}_{i{=}1}^n$ can be used to learn an inverse mapping $f\approx A^{-1}$ by minimizing the distance between $\x_i$ and $f(A \x_i)$, even when $A$ is underdetermined.
State-of-the-art methods usually parametrize the mapping functions with deep neural nets.
For example, stacked auto-encoders and convolutional neural nets have been used to solve compressive sensing and image deblurring problems~\cite{mousavi2015deep,kulkarni2016reconnet,xu2014deep,mousavi2017learning}.
%
%
Recently, adversarial learning~\cite{goodfellow2014generative} has been demonstrated its ability to solve many
challenging image problems, such as image inpainting~\cite{pathak2016context} and super-resolution~\cite{ledig2016photo,dahl2017pixel}.
%
%
%
%

Despite their ability to solve challenging problems, solving linear inverse problems with end-to-end mappings have a major disadvantage --- the number of mapping functions scales linearly with the number of problems.
Since the datasets are generated based on specific operators $A$s, these end-to-end mappings can only solve the given
problems $A$. 
Even if the problems change slightly, the mapping functions (nerual nets) need to be retrained.
For example, a mapping to solve $2\times$-super-resolution cannot be used directly to solve $4\times$-super-resolution with satisfactory performance; it is even more difficult to re-purpose a mapping for image inpainting to solve image super-resolution.
%
%
This specificity of end-to-end mappings makes it costly to incorporate them into consumer products that need to deal with a variety of image processing applications.

\myparagraph{Deep generative models.} Another thread of research learns generative models from image datasets.
Suppose we have a dataset containing samples of a distribution $P(\x)$. 
%
%
We can estimate $P(\x)$ and sample from the model~\cite{russ2007dbm, kingma2013vae,theis2015slstm}, or directly generate new samples from $P(\x)$ without explicitly estimating the distribution~\cite{goodfellow2014generative, radford2015unsupervised}.
Dave~\etal~\cite{dave2016compressive} use a spatial long-short-term memory network to learn the distribution $P(\x)$; to solve linear inverse problems, they solve a maximum a posteriori estimation --- maximizing $P(\x)$ for $\x$ such that $\y = A \x$.
Nguyen~\etal~\cite{nguyen2016plug} use a discriminative network and denoising autoencoders to implicitly learn the joint distribution between the image and its label $P(\x, y)$, and they generate new samples by sampling  the joint distribution $P(\x, y)$, \ie, the network, with an approximated Metropolis-adjusted Langevin algorithm.  
To solve image inpainting, they replace the values of known pixels in sampled images and  repeat the sampling process. 
Similar to the proposed framework, these methods can be used to solve a wide variety of inverse problems. 
They use a probability framework and thus can be considered orthogonal to the proposed framework, which is motivated by a geometric perspective.

\section{One Network to Solve Them All}
\label{sec: propose}

Signal priors play an important role in regularizing underdetermined inverse problems.
As mentioned in the introduction, traditional priors constraining the sparsity of signals in gradient or wavelet bases are often too generic, in that we can easily create non-image signals satisfying these priors.
Instead of using traditional signal priors, we propose to learn a prior from a large image dataset.
Since the  prior is learned directly from the dataset, it is tailored to the statistics of images in the dataset and, in principle, provide stronger regularization to the inverse problem.
In addition, similar to traditional signal priors, the learned signal prior can be used to solve any linear inverse problems pertaining to images.

\subsection{Problem formulation}

The proposed framework is motivated by the optimization technique, alternating direction method of multipliers method (ADMM)~\cite{boyd2011distributed}, that is widely used to solve linear inverse problems as defined in (\ref{eq: inverse}).
A typical first step in ADMM is to separate a complicated objective into several simpler ones by variable splitting, \ie, introducing an additional variable $\z$ that is constrained to be equal to $\x$.
This gives us the following optimization problem:
\begin{align}
& \min_{\x, \z} \  \frac{1}{2} \norm{\y - A \z }_2^2 + \lambda \, \phi(\x) \\
& \ \ \textrm{s.t.} \ \ \x = \z, 
\label{eq: split}
\end{align}
which is equivalent to the original problem~\eqref{eq: inverse}.  
The scaled form of the augmented Lagrangian of~\eqref{eq: split} can be written as 
\begin{equation}
\mL(\x, \z, \u)  = \frac{1}{2} \| \y - A \z \|_2^2 + \lambda \, \phi(\x) + \frac{\rho}{2} \norm{ \x - \z + \u }^2_2,
\end{equation}
where $\rho > 0$ is the penalty parameter of the constraint $\x = \z$, and $\u$ is the dual variables divided by $\rho$.
By alternatively optimizing $\mL(\x, \z, \u)$ over $\x$, $\z$, and $\u$, ADMM is composed of the following procedures: 
\small{
\begin{align}
\iterkplus{\x} & \leftarrow  \argmin_\x  \frac{\rho}{2} \norm{\x - \iterk{\z} +  \iterk{\u } }_2^2  + \lambda \, \phi(\x)  \label{eq: x update} \\
\iterkplus{\z} & \leftarrow  \argmin_{\z}  \frac{1}{2} \norm{ \y - A \z }_2^2  {+}  \frac{\rho}{2} \norm{ \iterkplus{\x} {-} \z {+} \iterk{\u} }^2_2 \label{eq: z update} \\
\iterkplus{\u}& \leftarrow \iterk{\u} + \iterkplus{\x} - \iterkplus{\z}.
\end{align}
}
The update of $\z$~\eqref{eq: z update} is a least squares problem and can be solved efficiently via algorithms like conjugate gradient descent.  
The update of $\x$~\eqref{eq: x update} is the proximal operator of the signal prior $\phi$ with penalty $\frac{\rho}{\lambda}$, denoted as $\prox_{\phi, \frac{\rho}{\lambda}}(\bv)$, where $\bv {=} \z^{(k)} {-}  \bu^{(k)}$.
When the signal prior uses $\ell_1$-norm, the proximal operator is simply a soft-thresholding on $\bv$.  
Notice that the ADMM algorithm separates the signal prior $\phi$ from the linear operator $A$.  
This enables us to learn a signal prior that can be used with any linear operator.

\begin{figure}[t]
	\centering
		\includegraphics[width=\linewidth]{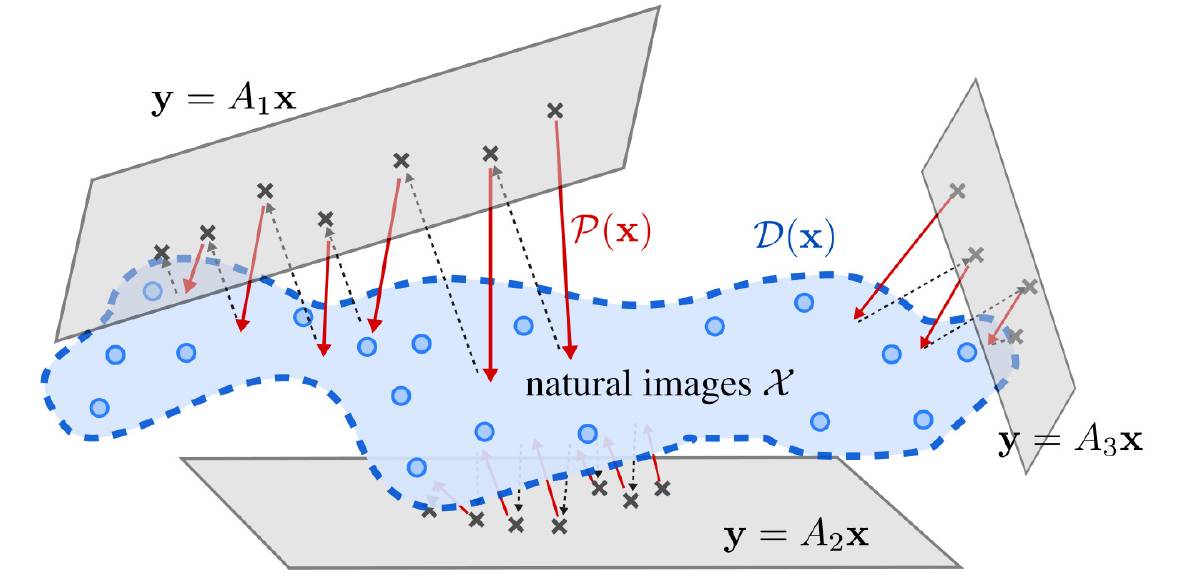}
		\caption{Given a large image dataset, the proposed framework learns a classifier $\mD$ that fits a decision boundary of the natural image set.  Based on $\mD$, a projection network $\mP(\x){:} \R^{d} {\rightarrow} \R^{d}$ is trained to fit the proximal operator of $\mD$, which enables to solve a variety of linear inverse problems related to images using ADMM.}
	\label{figure: geo}
\end{figure}

\subsection{Learning a proximal operator}

Since the signal prior only appears in the form of a proximal operator in ADMM, instead of explicitly learning a signal
prior $\phi$ and solving the proximal operator in each step of ADMM, we propose to directly learn the proximal operator.

Let $\cX$ represent the set of all natural images. 
The best signal prior is the indicator function of $\cX$, denoted as $\mI_{\cX}(\cdot)$, and its corresponding proximal operator $\prox_{\mI_{\cX}, \rho}(\bv)$ is a projection operator that projects $\bv$ onto $\cX$ from the geometric perspective--- or equivalently, finding a $\x \in \cX$ such that $\norm{\x - \bv}$ is minimized.
However, we do not have the oracle indicator function $\mI_{\cX}(\cdot)$ in practice, so we cannot evaluate $\prox_{\mI_{\cX}, \rho}(\bv)$ to solve the projection operation. 
Instead, we propose to train a classifier $\mD$ with a large dataset whose classification cost function approximates $\mI_{\cX}$. 
Based on the learned classifier $\mD$, we can learn a projection function $\cP$ that maps a signal $\bv$ to the set defined by the classifier.
The learned projection function $\cP$ can then replace the proximal operator~\eqref{eq: x update}, and we simply update $\x$ via
\begin{equation}
\iterkplus{\x} \leftarrow \mP(\iterk{\z} - \iterk{\u}).
\label{eq: new x update}
\end{equation} 
An illustration of the idea is shown in Figure~\ref{figure: geo}.

There are some caveats for this approach.  
First, when the classification cost function of the classifier $\mD$ is nonconvex, the overall optimization becomes nonconvex.
For solving general nonconvex optimization problems, the convergence result is not guaranteed. 
Based on the theorems for the convergence of nonconvex ADMM~\cite{wang2015global}, we provide the following theorem to the proposed ADMM framework. 
\begin{theorem}
Assume the  function $\mP$ solves the proximal operator~\eqref{eq: x update}. If the gradient of $\phi(x)$ is Lipschitz continuous and with large enough $\rho$, the ADMM algorithm is guaranteed to attain a stationary point.
\label{thm: conv}
\end{theorem}
The proof follows directly from~\cite{wang2015global} and we omit the details here.  
Although Theorem~\ref{thm: conv} only guarantees convergence to stationary points instead of the optimal solution as other nonconvex formulations, it ensures that the algorithm will not diverge after several iterations. 
Second, we initialize the scaled dual variables $\u$ with zeros and $\z^{(0)}$ with the pseudo-inverse of the least-square term.  
Since we initialize $\u^{0} = \mbf{0}$, the input to the proximal operator $\iterk{\bv} {=} \iterk{\z} {-} \iterk{\u} = \iterk{\z} + \sum_{i=1}^k \x^{(i)} - \z^{(i)} \approx \iterk{\z}$ resembles an image. 
Thereby, even though it is in general difficult to fit a projection function from any signal in $\R^d$ to the natural image space, we expect that the projection function only needs to deal with inputs that are close to images, and we train the projection function with slightly perturbed images from the dataset.
Third,  techniques like denoising autoencoders learn projection-like operators and, in principle, can be used  in place of a proximal operator; however,  our empirical findings suggest that  ignoring the projection cost $\| \bv - \cP(\bv) \|^2$ and simply minimizing the reconstruction loss $\| \x_0 - \cP(\bv) \|^2$, where $\bv$ is a perturbed image from $\x_0$, leads to instability in the ADMM iterations.

\subsection{Implementation details}
\label{sec: details}

We use two deep neural nets as the classifier $\mD$ and the projection operator $\mP$, respectively.
Based on Theorem~\ref{thm: conv}, we require the gradient of $\phi$ to be Lipschitz continuous.  
Since we choose to use cross entropy loss, we have $\phi(\x) = \log( \sigmoid(\mD(\x)))$ and in order to satisfy Theorem~\ref{thm: conv}, we need $\mD$ to be differentiable.
Thus, we use the smooth exponential linear unit~\cite{clevert2015fast} as activation function, instead of rectified linear units. 
To bound the gradients of $\mD$ \wrt $\x$, we truncate the weights of the network after each iteration.

We show an overview of the proposed method in Figure~\ref{figure: block diagram} and leave the details in Appendix~\ref{sec: arch}.  
The projector shares the same architecture of a typical convolutional autoencoder, and the classifier is a residual net~\cite{he2016deep}. 
One way to train the classifier $\mD$ is to feed $\mD$ natural images from a dataset and their perturbed counterparts.
Nevertheless, we expect the projected images produced by the projector $\mP$ be closer to the dataset $\mM$ (natural images) than those perturbed images.
Therefore, we jointly train two networks using adversarial learning:  
The projector $\mP$ is trained to minimize~\eqref{eq: x update}, that is, confusing the classifier $\mD$ by projecting $\bv$ to the natural image set defined by the decision boundary of $\mD$.
When the projector improves and generates outputs that are within or closer to the boundary, the classifier can be updated to tighten its decision boundary.
Although we start from a different perspective from~\cite{goodfellow2014generative}, the above joint training procedure can also be understood as a two player game in adversarial learning, where the projector $\mP$ tries to confuse the classifier $\mD$.

Specifically, we optimize the projection network with the following objective function: 
{\footnotesize 
\begin{align}
& \min_{\btheta_\cP} \sum_{\x \in \cM, \bv \sim f(\x)} \lambda_1  \| \x - \cP(\x) \|^2 
+\lambda_2  \| \x - \cP(\bv) \|^2  \label{eq: l2} \\ 
&  
 + \lambda_3  \| \bv - \cP(\bv)\|^2  - \lambda_4 \log\pqty{\sigmoid\pqty{\mD_\ell \circ \mE(\bv) } } - \lambda_5 \log\pqty{\sigmoid\pqty{\mD \circ \cP(\bv) }
\label{eq: adv}
}, 
\end{align}
}
where $f$ is the function we used to generate perturbed images, and  the first two terms in \eqref{eq: l2} are similar to (denoising) autoencoders and are added to help the training procedure.
The remaining terms in \eqref{eq: adv} form the projection loss as we need in~\eqref{eq: x update}. 
We use two classifiers $\mD$ and $\mD_\ell$, for the output (image) space and the latent spaces of the projector ($\mE(\bv)$ in Figure~\ref{figure: block diagram}), respectively. 
The latent classifier $\mD_\ell$ is added to further help the training procedure~\cite{Makhzani2015aae}. 
We find that adding $\mD_\ell$ also helps the projector to avoid overfitting.   
In all of our experiments, we set $\lambda_1 = 0.01, \lambda_3 = 0.005$, $\lambda_2 = 1.0$, $\lambda_4 = 0.0001$, and $\lambda_5 = 0.001$.
The diagram of the training objective is shown in~\ref{figure: block diagram}.

We briefly discuss the architecture of the networks.   
More details can be seen in Appendix~\ref{sec: arch}.
As we have discussed above, to improve the convergence of ADMM, we use exponential linear unit and truncate the weights of $\mD$ and $\mD_\ell$ after each training iteration. 
Following the guidelines in~\cite{radford2015unsupervised,salimans2016improved}, we use a $10$-layer convolution neural net that uses strides to downsample/upsample the images, and we use virtual batch normalization.  
However, we do not truncate the outputs of $\cP$ with a tanh or a sigmoid function, in order to authentically compute the $\ell_2$ projection loss.  
We find that using linear output helps the convergence of the ADMM procedure.
We also use residual nets with six residual blocks as $\mD$ and $\mD_\ell$ instead of typical convolutional neural nets.
We found that the stronger gradients provided by the shortcuts usually helps speed up the training process. 
Besides, we add a channel-wise fully connected layer followed by a $2\times 2$ convolution to enable the projector to learn the context in the image, as in~\cite{pathak2016context}.
The classifiers $\mD$ and $\mD_\ell$ are trained to minimize the negative cross entropy loss.
We use early stopping in order to alleviate overfitting.
The complete architecture information is shown in the supplemental material. 

\myparagraph{Image perturbation.}
While adding Gaussian noise may be the simplest method to perturb an image,  we found that the projection network will easily overfit the Gaussian noise and become a dedicated Gaussian denoiser.  
Since during the ADMM process, the inputs to the projection network, $\iterk{\z} - \iterk{\u}$, do not usually follow a Gaussian distribution, the overfitted projection network may fail to project the general signals produced by the ADMM process.
To avoid overfitting, we generate perturbed images with two methods --- adding Gaussian noise with spatially varying standard deviations and smoothing the input images.
We generate the noise by multiplying a randomly sampled standard Gaussian noise with a weighted mask upsampled from a low-dimensional mask with bicubic algorithm. 
The weighted mask is randomly sampled from a uniform distribution ranging from $[0.05, 0.5]$.  Note that the images are ranging from $[-1, 1]$. 
To smooth the input images, we first downsample the input images and then use nearest-neighbor method to upsample the results.  The ratio to the downsample is uniformly sampled from $[0.2, 0.95]$. 
After smoothing the images, we add the noise described above.
We only use the smoothed images on ImageNet and MS-Cele-1M datasets.

\begin{figure}[t]
	\centering
	\includegraphics[width=0.85\linewidth]{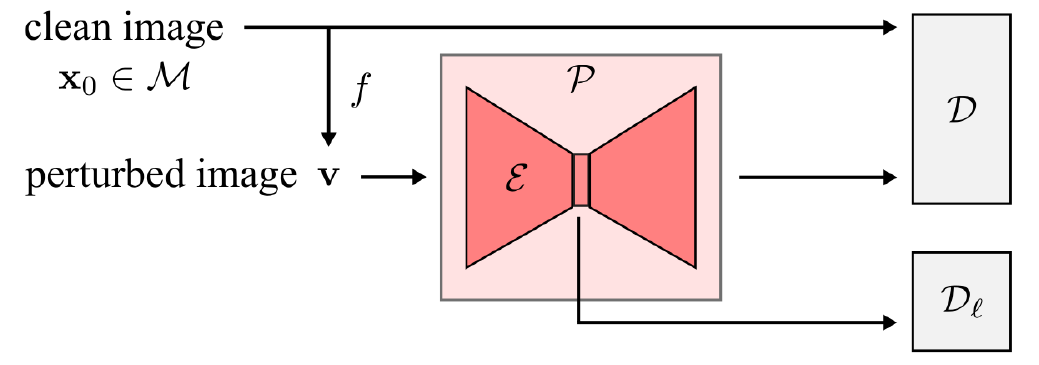}
	\caption{Training the projection network $\mP$.  The adversarial learning is conducted on both image and latent spaces of $\mP$. }
	\label{figure: block diagram}
\end{figure}

\def\dataset{venice_cropped_512}
\def\denoise{denoise_ratio0.50_std0.00}
\def\denoiseours{ours_alpha0.300000}
\def\cs{cs_ratio0.10_std0.00}
\def\csours{ours_alpha0.300000}
\def\inpaint{inpaint_bs6_tb400_std0.00}
\def\inpaintours{ours_alpha0.050000}
\def\superres{superres_ratio0.50_std0.00}
\def\superresours{ours_alpha0.500000}

\begin{figure*}[t]
	\centering
	\begin{subfigure}[t]{0.8\linewidth}
		\centering
		\hspace{-7.5mm}
		\begin{subfigure}[t]{0.48\linewidth}
			\centering
			\includegraphics[width=0.85\linewidth]{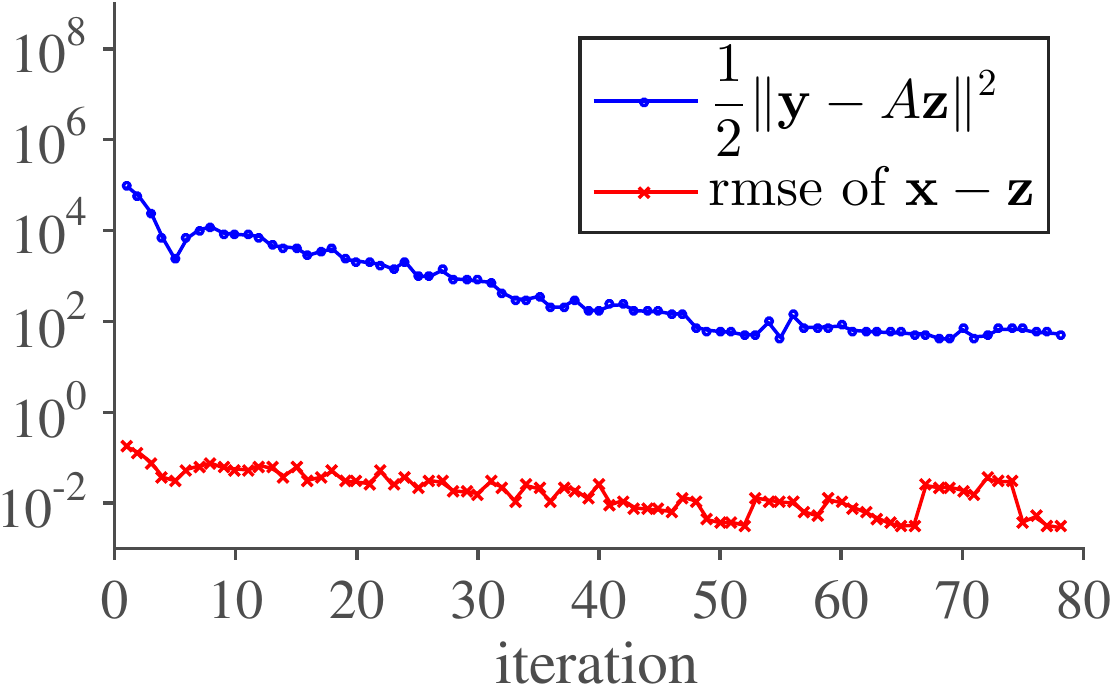}
		\end{subfigure}	
		\begin{subfigure}[t]{0.48\linewidth}
			\centering
			\includegraphics[width=0.85\linewidth]{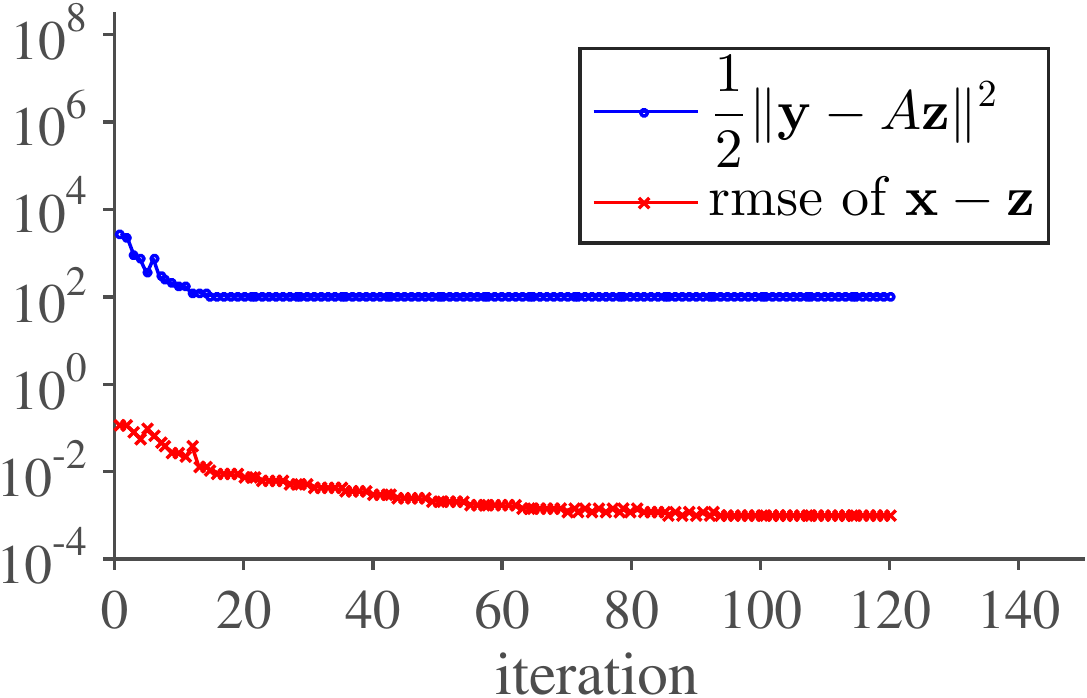}
		\end{subfigure}	
		\vspace{-1mm}
		\caption{ Iterations of ADMM for compressive sensing (left) and scattered inpainting (right) of results shown in Figure~\ref{figure: venice}. }
		\label{figure: convergence}
	\end{subfigure}
	\\
	\vspace{2mm}
	\begin{subfigure}[t]{0.8\linewidth}
		\begin{subfigure}{0.48\linewidth}
			\centering
			\includegraphics[width=\linewidth]{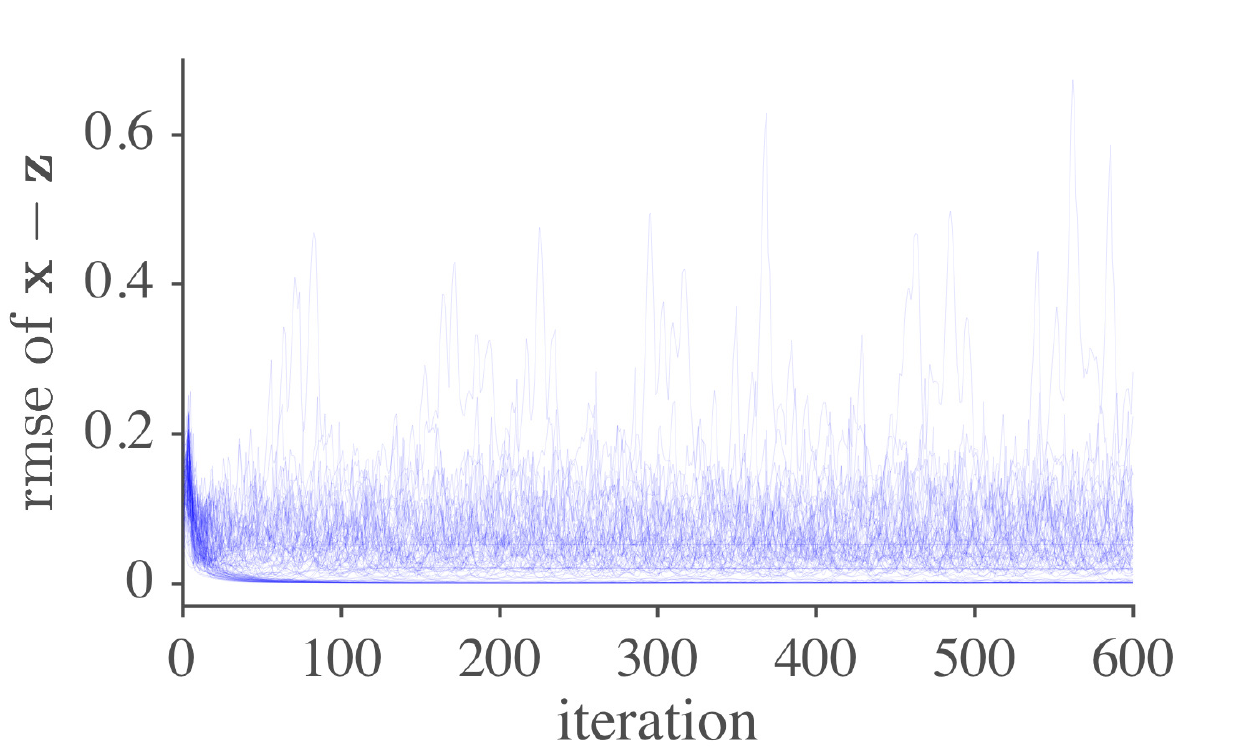}
			\vspace{-47mm}
			\caption*{\hspace{2mm} \scriptsize non-smooth $\mD$ and $\mD_\ell$}
		\end{subfigure}
		\begin{subfigure}{0.48\linewidth}
			\centering
			\includegraphics[width=\linewidth]{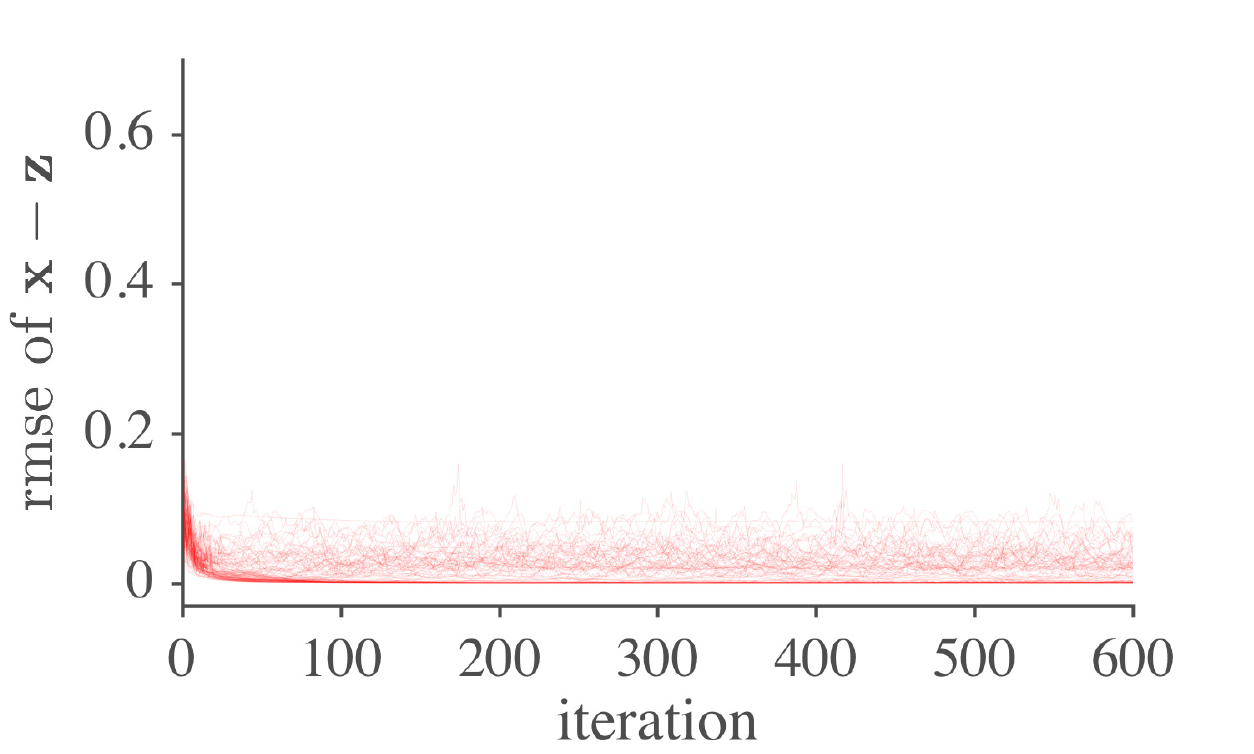}
			\vspace{-47mm}
			\caption*{\hspace{2mm} \scriptsize proposed architecture \\ \hspace{2mm} (smooth $\mD$ and $\mD_\ell$)}
		\end{subfigure}
		\vspace{35mm}
		\caption{Comparison of ADMM convergence between a projection network trained with indifferentiable $\mD$ and $\mD_\ell$ (left) and the proposed architecture (right), in which the gradient of $\mD$ and $\mD_\ell$ are Lipschitz continuous. We perform scattered inpainting with box size equal to $6$ and a total of $10$ boxes on $100$ random images in ImageNet dataset.   For both cases, we set $\rho = 0.05$.  We use transparent lines ($\alpha=0.1$) in order to show densities. }
		\label{figure: diff vs indiff}
	\end{subfigure}
	\caption{Convergence of ADMM}
\end{figure*}

\subsection{Relationship to other algorithms}

The proposed framework, which is composed of a classifier network and a projection network, is very similar to the works involving both adversarial learning and denoising auto-encoder, \eg, context encoder~\cite{pathak2016context} and mode-regularized GAN~\cite{che2016mode}.
Compared to the probabilistic perspective typically used in adversarial learning~\cite{goodfellow2014generative} that matches the distributions of the dataset and the generated images, the proposed framework is based on the geometric perspective and the ADMM framework. 
We approximate the oracle indicator function $\mI_{\cX}$ by $\mD$ and its proximal operator by the projection operator $\mP$. 
Our use of the adversarial training is simply for learning a tighter decision boundary, based on the hypothesis that images generated by $\mP$ should be closer in $\ell_2$ distance to $\cX$ than the arbitrarily perturbed images.

The projection network $\mP$ can be thought as a denoising autoencoder with a regularization term.  
Compared to the typical denoising auto-encoder, which always projects a perturbed image $\x_0 + \n$ back to the origin image $\x_0$, the proposed $\mP$ may project $\x_0 + \n$ to the closest $\x$ in $\cX$.  
In our empirical experience, the additional projection $\ell_2$ loss help stabilize the ADMM process.

A recent work by Dave~\etal can also be used to solve generic linear inverse problems.
They use a spatial recurrent generative network to model the distribution of natural images $P(\x) $and solve linear inverse problems by performing maximum a posteriori inference (maximizing $P(\x)$ given $\y = A \x$).
During the optimization, their method needs to compute the gradient of the network \wrt the input in each iteration, which can be computationally expensive when the network is very deep and complex.
In contrast, the proposed method directly provides the solution to the x-update~\eqref{eq: new x update} and is thus computationally efficient.

The proposed method is also very similar to the denoising-based approximate message passing algorithm (D-AMP)~\cite{metzler2016denoising}, which has achieved state-of-the-art performance in solving compressive sensing problems.
D-AMP is also motivated by geometric perspective of linear inverse problems and solve the compressive sensing problem with a variant of proximal gradient algorithm. 
However, instead of designing the proximal operator as the proposed method, D-AMP uses existing Gaussian denoising algorithms and relies on a Onsager correction term to ensure the noise resemble Gaussian noise. 
Thereby, D-AMP can only deal with linear operator $A$ formed by random Gaussian matrices.
In contrast, the proposed method directly learns a proximal operator that projects a signal to the image domain, and therefore, has fewer assumptions to the linear operator $A$.

\subsection{Limitations}

Unlike traditional signal priors whose weights $\lambda$ can be adjusted at the time of solving the optimization problem~\eqref{eq: inverse}, the prior weight of the proposed framework is fixed once the projection network is trained.   
While an ideal projection operator should not be affected by the value of the prior weights, sometimes, it may be preferable to control the effect of the signal prior to the solution.
In our experiments, we find that adjusting $\rho$ sometimes has similar effects as adjusting $\lambda$.

The convergence analysis of ADMM in Theorem~\ref*{thm: conv} is based on the assumption that the projection network can provide global optimum of~\eqref{eq: x update}.  However, in practice the optimality is not guaranteed.  While there are convergence analyses with inexact proximal operators, the general properties are too complicated to analyze for deep neural nets.   In practice, we find that for problems like pixelwise inpainting, compressive sensing, $2\times$ super-resolution and scattered inpainting the proposed framework converges gracefully, as shown in Figure~\ref{figure: convergence}, but for more challenging problems like image inpainting with large blocks and $4\times$-super-resolution on ImageNet dataset, we sometimes need to stop the ADMM procedure early.

\def\dataset{mnist}
\def\denoiseone{denoise_ratio0.50_std0.00}
\def\denoiseoneours{ours_alpha0.100000}
\def\denoiseonel{l1_lambdal10.030000_alpha0.300000}
\def\denoisetwo{denoise_ratio0.70_std0.30}
\def\denoisetwoours{ours_alpha0.100000}
\def\denoisetwol{l1_lambdal10.030000_alpha0.300000}
\def\csone{cs_ratio0.30_std0.00}
\def\csoneours{ours_alpha0.300000}
\def\csonel{l1_lambdal10.030000_alpha0.300000}
\def\cstwo{cs_ratio0.05_std0.00}
\def\cstwoours{ours_alpha0.300000}
\def\cstwol{l1_lambdal10.030000_alpha0.300000}
\def\inpaintc{inpaintcenter_bs9_std0.00}
\def\inpaintcours{ours_alpha0.050000}
\def\inpaintcl{l1_lambdal10.030000_alpha0.300000}
\def\superresone{superres_ratio0.50_std0.00}
\def\superresoneours{ours_alpha0.300000}
\def\superresonel{l1_lambdal10.030000_alpha0.300000}
\def\superrestwo{superres_ratio0.25_std0.00}
\def\superrestwoours{ours_alpha0.300000}
\def\superrestwol{l1_lambdal10.030000_alpha0.300000}

\begin{figure*}[thhh]
	\centering
	\def\idxone{5}
	\def\idxtwo{77}
	\def\idxthree{27}
	\def\idxfour{9}
	\begin{subfigure}{\linewidth}
		\centering
		\begin{subfigure}[t]{\fifteenimgwidth}
			\centering
			\caption*{\Cblankcol}
		\end{subfigure}	
		\begin{subfigure}[t]{\fifteenimgwidth}
			\centering
			\caption*{\Ccscol 1}
		\end{subfigure}	
		\begin{subfigure}[t]{\fifteenimgwidth}
			\centering
			\caption*{\Ccscol 2}
		\end{subfigure}	
		\begin{subfigure}[t]{\fifteenimgwidth}
			\centering
			\caption*{\Cdenoisecol 1}
		\end{subfigure}	
		\begin{subfigure}[t]{\fifteenimgwidth}
			\centering
			\caption*{\Cdenoisecol 2}
		\end{subfigure}	
		\begin{subfigure}[t]{\fifteenimgwidth}
			\centering
			\caption*{\Cblockcol}
		\end{subfigure}	
		\begin{subfigure}[t]{\fifteenimgwidth}
			\centering
			\caption*{\Csuperrescola}
		\end{subfigure}	
		\begin{subfigure}[t]{\fifteenimgwidth}
			\centering
			\caption*{\Csuperrescolb}
		\end{subfigure}	
	\hspace{0.5mm}
		\begin{subfigure}[t]{\fifteenimgwidth}
			\centering
			\caption*{\Ccscol 1}
		\end{subfigure}	
		\begin{subfigure}[t]{\fifteenimgwidth}
			\centering
			\caption*{\Ccscol 2}
		\end{subfigure}	
		\begin{subfigure}[t]{\fifteenimgwidth}
			\centering
			\caption*{\Cdenoisecol 1}
		\end{subfigure}	
		\begin{subfigure}[t]{\fifteenimgwidth}
			\centering
			\caption*{\Cdenoisecol 2}
		\end{subfigure}	
		\begin{subfigure}[t]{\fifteenimgwidth}
			\centering
			\caption*{\Cblockcol}
		\end{subfigure}	
		\begin{subfigure}[t]{\fifteenimgwidth}
			\centering
			\caption*{\Csuperrescola}
		\end{subfigure}	
		\begin{subfigure}[t]{\fifteenimgwidth}
			\centering
			\caption*{\Csuperrescolb}
		\end{subfigure}	
	\end{subfigure}
	\\
	\vspace{\fifteenimgrowmargin}
	\begin{subfigure}{\linewidth}
		\centering
		\begin{subfigure}[t]{\fifteenimgwidth}
			\centering
			\vspace{\fifteenimgrowcaptionheight}
			\caption*{\groundtruthrowcaption}
		\end{subfigure}	
		\begin{subfigure}[t]{\fifteenimgwidth}
			\centering
			\includegraphics[width=\textwidth]{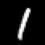}
		\end{subfigure}	
		\begin{subfigure}[t]{\fifteenimgwidth}
			\centering
			\includegraphics[width=\textwidth]{figures/\dataset/\idxone/ori_img.jpg}
		\end{subfigure}	
		\begin{subfigure}[t]{\fifteenimgwidth}
			\centering
			\includegraphics[width=\textwidth]{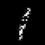}
		\end{subfigure}	
		\begin{subfigure}[t]{\fifteenimgwidth}
			\centering
			\includegraphics[width=\textwidth]{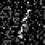}
		\end{subfigure}	
		\begin{subfigure}[t]{\fifteenimgwidth}
			\centering
			\includegraphics[width=\textwidth]{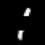}
		\end{subfigure}	
		\begin{subfigure}[t]{\fifteenimgwidth}
			\centering
			\vspace{-0.75\textwidth}
			\includegraphics[width=0.5\textwidth]{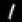}
		\end{subfigure}	
		\begin{subfigure}[t]{\fifteenimgwidth}
			\centering
			\vspace{-0.625\textwidth}
			\includegraphics[width=0.25\textwidth]{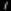}
		\end{subfigure}	
		\hspace{0.5mm}
		\begin{subfigure}[t]{\fifteenimgwidth}
			\centering
			\includegraphics[width=\textwidth]{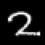}
		\end{subfigure}	
		\begin{subfigure}[t]{\fifteenimgwidth}
			\centering
			\includegraphics[width=\textwidth]{figures/\dataset/\idxtwo/ori_img.jpg}
		\end{subfigure}	
		\begin{subfigure}[t]{\fifteenimgwidth}
			\centering
			\includegraphics[width=\textwidth]{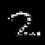}
		\end{subfigure}	
		\begin{subfigure}[t]{\fifteenimgwidth}
			\centering
			\includegraphics[width=\textwidth]{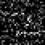}
		\end{subfigure}	
		\begin{subfigure}[t]{\fifteenimgwidth}
			\centering
			\includegraphics[width=\textwidth]{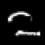}
		\end{subfigure}	
		\begin{subfigure}[t]{\fifteenimgwidth}
			\centering
			\vspace{-0.75\textwidth}
			\includegraphics[width=0.5\textwidth]{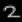}
		\end{subfigure}	
		\begin{subfigure}[t]{\fifteenimgwidth}
			\centering
			\vspace{-0.625\textwidth}
			\includegraphics[width=0.25\textwidth]{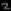}
		\end{subfigure}	
	\end{subfigure}
	\\
	\vspace{-1mm}
	\begin{subfigure}{\linewidth}
		\centering
		\begin{subfigure}[t]{\fifteenimgwidth}
			\centering
			\vspace{\fifteenimgrowcaptionheight}
			\vspace{3mm}
			\caption*{\oursrowcaption}
		\end{subfigure}	
		\begin{subfigure}[t]{\fifteenimgwidth}
			\centering
			\includegraphics[width=\textwidth]{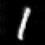}
		\end{subfigure}	
		\begin{subfigure}[t]{\fifteenimgwidth}
			\centering
			\includegraphics[width=\textwidth]{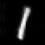}
		\end{subfigure}	
		\begin{subfigure}[t]{\fifteenimgwidth}
			\centering
			\includegraphics[width=\textwidth]{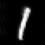}
		\end{subfigure}	
		\begin{subfigure}[t]{\fifteenimgwidth}
			\centering
			\includegraphics[width=\textwidth]{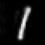}
		\end{subfigure}	
		\begin{subfigure}[t]{\fifteenimgwidth}
			\centering
			\includegraphics[width=\textwidth]{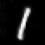}
		\end{subfigure}	
		\begin{subfigure}[t]{\fifteenimgwidth}
			\centering
			\includegraphics[width=\textwidth]{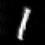}
		\end{subfigure}	
		\begin{subfigure}[t]{\fifteenimgwidth}
			\centering
			\includegraphics[width=\textwidth]{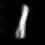}
		\end{subfigure}	
		\hspace{0.5mm}
		\begin{subfigure}[t]{\fifteenimgwidth}
			\centering
			\includegraphics[width=\textwidth]{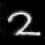}
		\end{subfigure}	
		\begin{subfigure}[t]{\fifteenimgwidth}
			\centering
			\includegraphics[width=\textwidth]{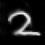}
		\end{subfigure}	
		\begin{subfigure}[t]{\fifteenimgwidth}
			\centering
			\includegraphics[width=\textwidth]{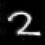}
		\end{subfigure}	
		\begin{subfigure}[t]{\fifteenimgwidth}
			\centering
			\includegraphics[width=\textwidth]{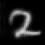}
		\end{subfigure}	
		\begin{subfigure}[t]{\fifteenimgwidth}
			\centering
			\includegraphics[width=\textwidth]{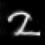}
		\end{subfigure}	
		\begin{subfigure}[t]{\fifteenimgwidth}
			\centering
			\includegraphics[width=\textwidth]{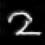}
		\end{subfigure}	
		\begin{subfigure}[t]{\fifteenimgwidth}
			\centering
			\includegraphics[width=\textwidth]{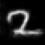}
		\end{subfigure}	
	\end{subfigure}
	\\
	\vspace{0.3mm}
	\begin{subfigure}{\linewidth}
		\centering
		\begin{subfigure}[t]{\fifteenimgwidth}
			\centering
			\vspace{\fifteenimgrowcaptionheight}
			\vspace{3mm}
			\caption*{\lrowcaption}
		\end{subfigure}	
		\hspace{0.02mm}
		\begin{subfigure}[t]{\fifteenimgwidth}
			\centering
			\includegraphics[width=\textwidth]{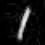}
		\end{subfigure}	
		\begin{subfigure}[t]{\fifteenimgwidth}
			\centering
			\includegraphics[width=\textwidth]{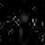}
		\end{subfigure}	
		\begin{subfigure}[t]{\fifteenimgwidth}
			\centering
			\includegraphics[width=\textwidth]{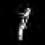}
		\end{subfigure}	
		\begin{subfigure}[t]{\fifteenimgwidth}
			\centering
			\includegraphics[width=\textwidth]{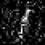}
		\end{subfigure}	
		\begin{subfigure}[t]{\fifteenimgwidth}
			\centering
			\includegraphics[width=\textwidth]{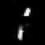}
		\end{subfigure}	
		\begin{subfigure}[t]{\fifteenimgwidth}
			\centering
			\includegraphics[width=\textwidth]{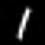}
		\end{subfigure}	
		\begin{subfigure}[t]{\fifteenimgwidth}
			\centering
			\includegraphics[width=\textwidth]{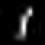}
		\end{subfigure}	
		\hspace{0.5mm}
		\begin{subfigure}[t]{\fifteenimgwidth}
			\centering
			\includegraphics[width=\textwidth]{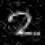}
		\end{subfigure}	
		\begin{subfigure}[t]{\fifteenimgwidth}
			\centering
			\includegraphics[width=\textwidth]{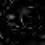}
		\end{subfigure}	
		\begin{subfigure}[t]{\fifteenimgwidth}
			\centering
			\includegraphics[width=\textwidth]{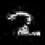}
		\end{subfigure}	
		\begin{subfigure}[t]{\fifteenimgwidth}
			\centering
			\includegraphics[width=\textwidth]{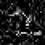}
		\end{subfigure}	
		\begin{subfigure}[t]{\fifteenimgwidth}
			\centering
			\includegraphics[width=\textwidth]{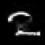}
		\end{subfigure}	
		\begin{subfigure}[t]{\fifteenimgwidth}
			\centering
			\includegraphics[width=\textwidth]{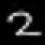}
		\end{subfigure}	
		\begin{subfigure}[t]{\fifteenimgwidth}
			\centering
			\includegraphics[width=\textwidth]{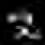}
		\end{subfigure}	
		\vspace{1mm}
	\end{subfigure}
	\def\idxone{\idxthree}
	\def\idxtwo{\idxfour}
	\begin{subfigure}{\linewidth}
		\centering
		\begin{subfigure}[t]{\fifteenimgwidth}
			\centering
			\vspace{\fifteenimgrowcaptionheight}
			\caption*{\groundtruthrowcaption}
		\end{subfigure}	
		\begin{subfigure}[t]{\fifteenimgwidth}
			\centering
			\includegraphics[width=\textwidth]{figures/\dataset/\idxone/ori_img.jpg}
		\end{subfigure}	
		\begin{subfigure}[t]{\fifteenimgwidth}
			\centering
			\includegraphics[width=\textwidth]{figures/\dataset/\idxone/ori_img.jpg}
		\end{subfigure}	
		\begin{subfigure}[t]{\fifteenimgwidth}
			\centering
			\includegraphics[width=\textwidth]{figures/\dataset/\idxone/\denoiseone/y.jpg}
		\end{subfigure}	
		\begin{subfigure}[t]{\fifteenimgwidth}
			\centering
			\includegraphics[width=\textwidth]{figures/\dataset/\idxone/\denoisetwo/y.jpg}
		\end{subfigure}	
		\begin{subfigure}[t]{\fifteenimgwidth}
			\centering
			\includegraphics[width=\textwidth]{figures/\dataset/\idxone/\inpaintc/y.jpg}
		\end{subfigure}	
		\begin{subfigure}[t]{\fifteenimgwidth}
			\centering
			\vspace{-0.75\textwidth}
			\includegraphics[width=0.5\textwidth]{figures/{\dataset}/{\idxone}/{\superresone}/y.jpg}
		\end{subfigure}	
		\begin{subfigure}[t]{\fifteenimgwidth}
			\centering
			\vspace{-0.625\textwidth}
			\includegraphics[width=0.25\textwidth]{figures/{\dataset}/{\idxone}/{\superrestwo}/y.jpg}
		\end{subfigure}	
		\hspace{0.5mm}
		\begin{subfigure}[t]{\fifteenimgwidth}
			\centering
			\includegraphics[width=\textwidth]{figures/\dataset/\idxtwo/ori_img.jpg}
		\end{subfigure}	
		\begin{subfigure}[t]{\fifteenimgwidth}
			\centering
			\includegraphics[width=\textwidth]{figures/\dataset/\idxtwo/ori_img.jpg}
		\end{subfigure}	
		\begin{subfigure}[t]{\fifteenimgwidth}
			\centering
			\includegraphics[width=\textwidth]{figures/\dataset/\idxtwo/\denoiseone/y.jpg}
		\end{subfigure}	
		\begin{subfigure}[t]{\fifteenimgwidth}
			\centering
			\includegraphics[width=\textwidth]{figures/\dataset/\idxtwo/\denoisetwo/y.jpg}
		\end{subfigure}	
		\begin{subfigure}[t]{\fifteenimgwidth}
			\centering
			\includegraphics[width=\textwidth]{figures/\dataset/\idxtwo/\inpaintc/y.jpg}
		\end{subfigure}	
		\begin{subfigure}[t]{\fifteenimgwidth}
			\centering
			\vspace{-0.75\textwidth}
			\includegraphics[width=0.5\textwidth]{figures/{\dataset}/{\idxtwo}/{\superresone}/y.jpg}
		\end{subfigure}	
		\begin{subfigure}[t]{\fifteenimgwidth}
			\centering
			\vspace{-0.625\textwidth}
			\includegraphics[width=0.25\textwidth]{figures/{\dataset}/{\idxtwo}/{\superrestwo}/y.jpg}
		\end{subfigure}	
	\end{subfigure}
	\\
	\vspace{-1mm}
	\begin{subfigure}{\linewidth}
		\centering
		\begin{subfigure}[t]{\fifteenimgwidth}
			\centering
			\vspace{\fifteenimgrowcaptionheight}
			\vspace{3mm}
			\caption*{\oursrowcaption}
		\end{subfigure}	
		\begin{subfigure}[t]{\fifteenimgwidth}
			\centering
			\includegraphics[width=\textwidth]{figures/\dataset/\idxone/\csone/\csoneours/z.jpg}
		\end{subfigure}	
		\begin{subfigure}[t]{\fifteenimgwidth}
			\centering
			\includegraphics[width=\textwidth]{figures/\dataset/\idxone/\cstwo/\cstwoours/z.jpg}
		\end{subfigure}	
		\begin{subfigure}[t]{\fifteenimgwidth}
			\centering
			\includegraphics[width=\textwidth]{figures/\dataset/\idxone/\denoiseone/\denoiseoneours/z.jpg}
		\end{subfigure}	
		\begin{subfigure}[t]{\fifteenimgwidth}
			\centering
			\includegraphics[width=\textwidth]{figures/\dataset/\idxone/\denoisetwo/\denoisetwoours/z.jpg}
		\end{subfigure}	
		\begin{subfigure}[t]{\fifteenimgwidth}
			\centering
			\includegraphics[width=\textwidth]{figures/\dataset/\idxone/\inpaintc/\inpaintcours/z.jpg}
		\end{subfigure}	
		\begin{subfigure}[t]{\fifteenimgwidth}
			\centering
			\includegraphics[width=\textwidth]{figures/{\dataset}/{\idxone}/{\superresone}/\superresoneours/z.jpg}
		\end{subfigure}	
		\begin{subfigure}[t]{\fifteenimgwidth}
			\centering
			\includegraphics[width=\textwidth]{figures/{\dataset}/{\idxone}/{\superrestwo}/\superrestwoours/z.jpg}
		\end{subfigure}	
		\hspace{0.5mm}
		\begin{subfigure}[t]{\fifteenimgwidth}
			\centering
			\includegraphics[width=\textwidth]{figures/\dataset/\idxtwo/\csone/\csoneours/z.jpg}
		\end{subfigure}	
		\begin{subfigure}[t]{\fifteenimgwidth}
			\centering
			\includegraphics[width=\textwidth]{figures/\dataset/\idxtwo/\cstwo/\cstwoours/z.jpg}
		\end{subfigure}	
		\begin{subfigure}[t]{\fifteenimgwidth}
			\centering
			\includegraphics[width=\textwidth]{figures/\dataset/\idxtwo/\denoiseone/\denoiseoneours/z.jpg}
		\end{subfigure}	
		\begin{subfigure}[t]{\fifteenimgwidth}
			\centering
			\includegraphics[width=\textwidth]{figures/\dataset/\idxtwo/\denoisetwo/\denoisetwoours/z.jpg}
		\end{subfigure}	
		\begin{subfigure}[t]{\fifteenimgwidth}
			\centering
			\includegraphics[width=\textwidth]{figures/\dataset/\idxtwo/\inpaintc/\inpaintcours/z.jpg}
		\end{subfigure}	
		\begin{subfigure}[t]{\fifteenimgwidth}
			\centering
			\includegraphics[width=\textwidth]{figures/{\dataset}/{\idxtwo}/{\superresone}/\superresoneours/z.jpg}
		\end{subfigure}	
		\begin{subfigure}[t]{\fifteenimgwidth}
			\centering
			\includegraphics[width=\textwidth]{figures/{\dataset}/{\idxtwo}/{\superrestwo}/\superrestwoours/z.jpg}
		\end{subfigure}	
	\end{subfigure}
	\\
	\vspace{0.3mm}
	\begin{subfigure}{\linewidth}
		\centering
		\begin{subfigure}[t]{\fifteenimgwidth}
			\centering
			\vspace{\fifteenimgrowcaptionheight}
			\vspace{3mm}
			\caption*{\lrowcaption}
		\end{subfigure}	
		\begin{subfigure}[t]{\fifteenimgwidth}
			\centering
			\includegraphics[width=\textwidth]{figures/\dataset/\idxone/\csone/\csonel/z.jpg}
		\end{subfigure}	
		\begin{subfigure}[t]{\fifteenimgwidth}
			\centering
			\includegraphics[width=\textwidth]{figures/\dataset/\idxone/\cstwo/\cstwol/z.jpg}
		\end{subfigure}	
		\begin{subfigure}[t]{\fifteenimgwidth}
			\centering
			\includegraphics[width=\textwidth]{figures/\dataset/\idxone/\denoiseone/\denoiseonel/z.jpg}
		\end{subfigure}	
		\begin{subfigure}[t]{\fifteenimgwidth}
			\centering
			\includegraphics[width=\textwidth]{figures/\dataset/\idxone/\denoisetwo/\denoisetwol/z.jpg}
		\end{subfigure}	
		\begin{subfigure}[t]{\fifteenimgwidth}
			\centering
			\includegraphics[width=\textwidth]{figures/\dataset/\idxone/\inpaintc/\inpaintcl/z.jpg}
		\end{subfigure}	
		\begin{subfigure}[t]{\fifteenimgwidth}
			\centering
			\includegraphics[width=\textwidth]{figures/{\dataset}/{\idxone}/{\superresone}/\superresonel/z.jpg}
		\end{subfigure}	
		\begin{subfigure}[t]{\fifteenimgwidth}
			\centering
			\includegraphics[width=\textwidth]{figures/{\dataset}/{\idxone}/{\superrestwo}/\superrestwol/z.jpg}
		\end{subfigure}	
		\hspace{0.5mm}
		\begin{subfigure}[t]{\fifteenimgwidth}
			\centering
			\includegraphics[width=\textwidth]{figures/\dataset/\idxtwo/\csone/\csonel/z.jpg}
		\end{subfigure}	
		\begin{subfigure}[t]{\fifteenimgwidth}
			\centering
			\includegraphics[width=\textwidth]{figures/\dataset/\idxtwo/\cstwo/\cstwol/z.jpg}
		\end{subfigure}	
		\begin{subfigure}[t]{\fifteenimgwidth}
			\centering
			\includegraphics[width=\textwidth]{figures/\dataset/\idxtwo/\denoiseone/\denoiseonel/z.jpg}
		\end{subfigure}	
		\begin{subfigure}[t]{\fifteenimgwidth}
			\centering
			\includegraphics[width=\textwidth]{figures/\dataset/\idxtwo/\denoisetwo/\denoisetwol/z.jpg}
		\end{subfigure}	
		\begin{subfigure}[t]{\fifteenimgwidth}
			\centering
			\includegraphics[width=\textwidth]{figures/\dataset/\idxtwo/\inpaintc/\inpaintcl/z.jpg}
		\end{subfigure}	
		\begin{subfigure}[t]{\fifteenimgwidth}
			\centering
			\includegraphics[width=\textwidth]{figures/{\dataset}/{\idxtwo}/{\superresone}/\superresonel/z.jpg}
		\end{subfigure}	
		\begin{subfigure}[t]{\fifteenimgwidth}
			\centering
			\includegraphics[width=\textwidth]{figures/{\dataset}/{\idxtwo}/{\superrestwo}/\superrestwol/z.jpg}
		\end{subfigure}	
	\end{subfigure}
	\vspace{-1mm}
	\caption{ Results on MNIST dataset.  Since the input of compressive sensing cannot be visualized, we show the ground truth instead.  Compressive sensing~1 uses $\frac{m}{d}=0.3$ and compressive sensing~2 uses $\frac{m}{d}=0.03.$ Pixelwise inpaint~1 drops $50\%$ of the pixels, and pixelwise inpaint~2 drops $70\%$ of the pixels and add Gaussian noise with $\sigma=0.3$.  We use $\rho=0.1$ for pixelwise inpainting and $\rho=0.05$ for blockwise inpainting.}
	\label{figure: mnist}
	\vspace{-2mm}
\end{figure*}

\section{Experiments}

We evaluate the proposed framework on three datasets: 
\begin{description}[align=left,style=unboxed,leftmargin=2mm]
	\item [MNIST dataset~\cite{loosli-canu-bottou-2006}] contains randomly deformed images of MNIST dataset. The images are $28\times 28$ and grayscale. We train the projector and the classifier networks on the training set and test the results on the test set.  Since the dataset is relatively simpler, we remove the upper three layers from both $\mD$ and $\mD_\ell$.  We use batches with $32$ instances and train the networks for $80000$ iterations.
	\item [MS-Celeb-1M dataset~\cite{guo2016ms}] contains a total of 8 million aligned and cropped face images of 10 thousand people from different viewing angles.  We randomly select images of 73678 people as the training set and those of 25923 people as the test set. We resize the images into $64 \times 64$.  We use batches with $25$ instances and train the network for $10000$ iterations. 
	\item [ImageNet dataset~\cite{russakovsky2015imagenet}] contains 1.2 million training images and 100 thousand test images on the Internet.  We resize the images into $64 \times 64$.  We use batches with $25$ instances and train the network for $68000$ iterations. 
\end{description}
For each of the datasets, we perform the following tasks: 
\begin{description}[align=left,style=unboxed,leftmargin=2mm]
	\item [Compressive sensing.] We use $m \times d$ random Gaussian matrices of different compression ($\frac{m}{d}$) as the linear operator $A$.  The images are vectorized into $d$-dimensional vectors $\x$ and multiplied with the random Gaussian matrices to form $\y$.
	\item [Pixelwise inpainting and denoising.] We randomly drop pixel values (independent of channels) by filling zeros and add Gaussian noise with different standard deviations.
	\item [Scattered inpainting.]  We randomly drop $10$ small blocks by filling zeros.  Each block is of $10\%$ width and height of the input.
	\item [Blockwise inpainting.]  We fill the center $30\%$ region of the input images with zeros. 
	\item [Super resolution.]  We downsample the images  into $50\%$ and $25\%$ of the original width and height using box-averaging algorithm.
\end{description}

\myparagraph{Comparison to specially-trained networks.}   
For each of the tasks, we train a specially-trained neural network using context encoder~\cite{pathak2016context} with adversarial training.
For compressive sensing, we design the network based on the work of~\cite{mousavi2017learning}, which applies $A^\top$ to the linear measurements and resize it into the image size to operate in image space.     
The measurement matrix $A$ is a random Gaussian matrix and is fixed. 
For pixelwise inpainting and denoise, we randomly drop $50\%$ of the pixels and add Gaussian noise with $\sigma=0.5$ for each training instances. 
For blockwise inpainting, we drop a block with $30\%$ size of the images at a random location in the images. 
For super resolution, we follow the work of Dong~\etal~\cite{dong2014learning} which first upsamples the low-resolution images to the target resolution using bicubic algorithm. 
We train a network for $2\times$-super resolution. 
Note that we do not train a network for $4\times$-super resolution and for scattered inpainting ---  to demonstrate that the specially-trained networks do not generalize well to similar tasks. 
Since the inputs to the $2\times$-super resolution network are bicubic-upsampled images, we also apply the upsampling to $\frac{1}{4}$-resolution images and feed them to the same network.
We also feed scattered inpainting inputs to the blockwise inpainting network.

\begin{figure*}[t]
	\centering
	\def\idx{2}
	\begin{subfigure}{0.24\linewidth}
		\centering
		\includegraphics[width=0.25\linewidth]{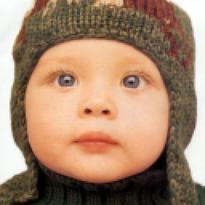}
		\vspace{-2mm}
		\caption*{\scriptsize input}
	\end{subfigure}
	\begin{subfigure}{0.24\linewidth}
		\centering
		\includegraphics[width=\linewidth]{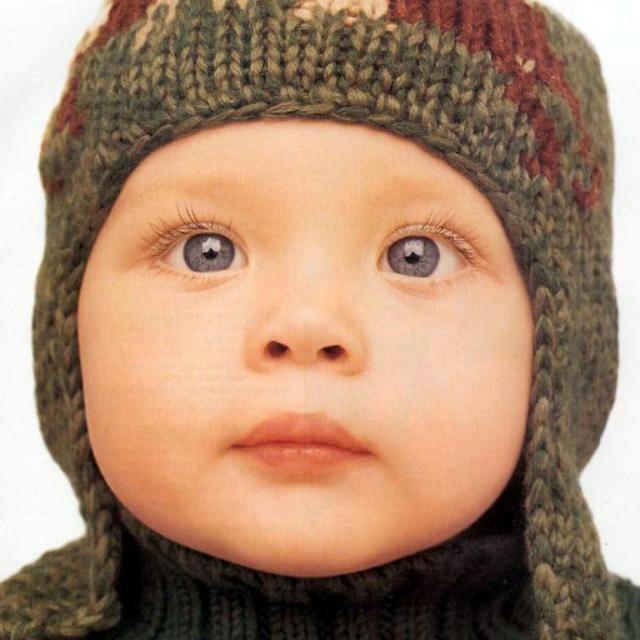}
		\vspace{-6mm}
		\caption*{\scriptsize ground truth}
	\end{subfigure}
	\begin{subfigure}{0.24\linewidth}
		\centering
		\includegraphics[width=\linewidth]{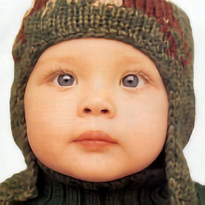}
		\vspace{-6mm}
		\caption*{\scriptsize \cite{FreFat10} (PSNR=$20.50$)}
	\end{subfigure}
	\begin{subfigure}{0.24\linewidth}
		\centering
		\includegraphics[width=\linewidth]{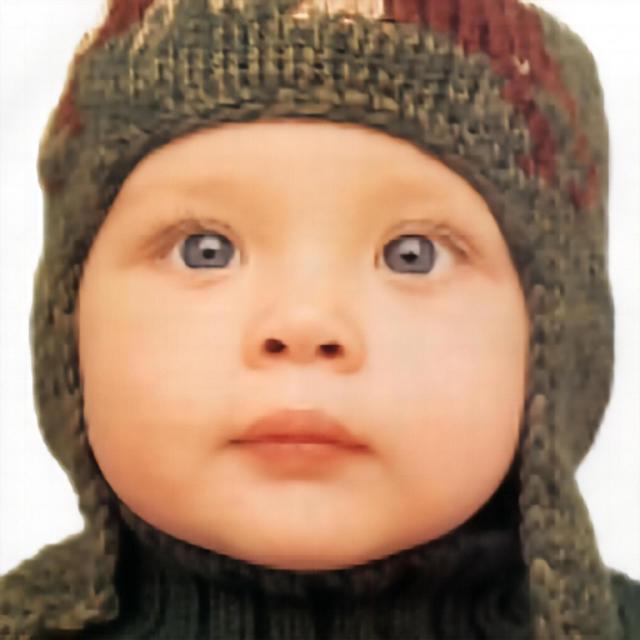}
		\vspace{-6mm}
		\caption*{\scriptsize proposed (PSNR=$23.20$)}
	\end{subfigure}
	\\
	\vspace{3mm}
	\def\idx{3}
	\begin{subfigure}{0.24\linewidth}
		\centering
		\includegraphics[width=0.25\linewidth]{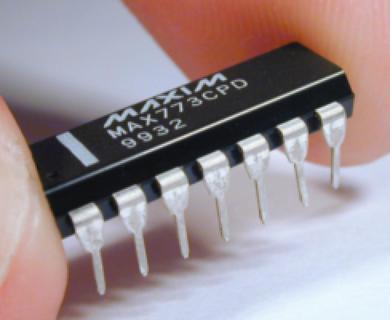}
		\vspace{-2mm}
		\caption*{\scriptsize input}
	\end{subfigure}
	\begin{subfigure}{0.24\linewidth}
		\centering
		\includegraphics[width=\linewidth]{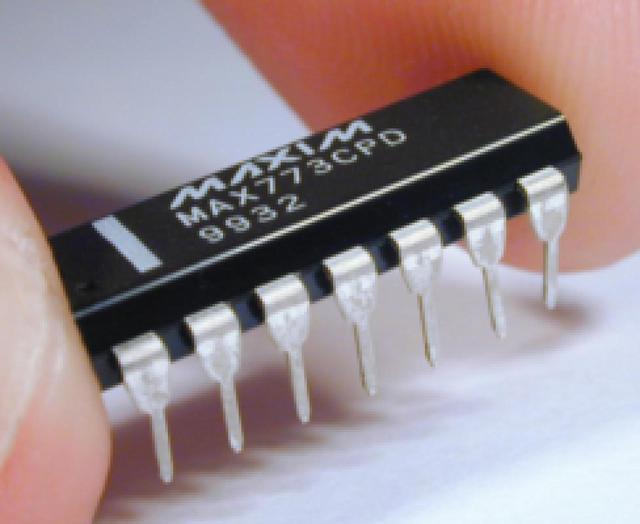}
		\vspace{-6mm}
		\caption*{\scriptsize bicubic interpolation}
	\end{subfigure}
	\begin{subfigure}{0.24\linewidth}
		\centering
		\includegraphics[width=\linewidth]{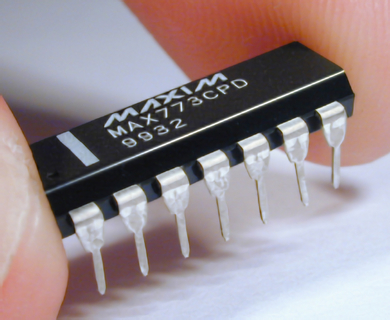}
		\vspace{-6mm}
		\caption*{\scriptsize \cite{FreFat10}}
	\end{subfigure}
	\begin{subfigure}{0.24\linewidth}
		\centering
		\includegraphics[width=\linewidth]{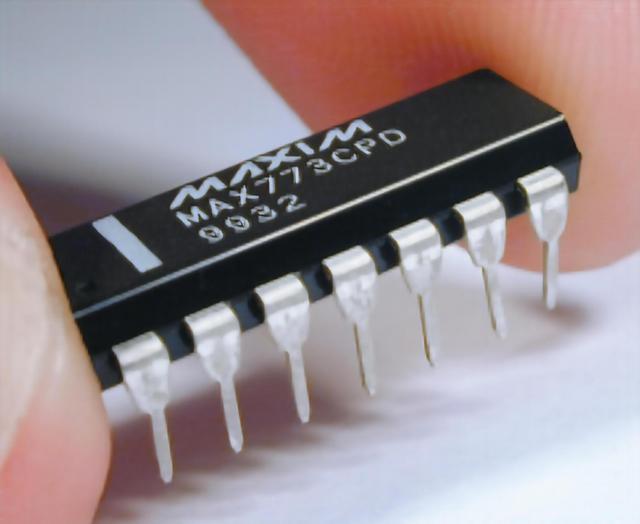}
		\vspace{-6mm}
		\caption*{\scriptsize proposed}
	\end{subfigure}
	\\
	\vspace{3mm}
	\def\idx{8}
	\begin{subfigure}{0.24\linewidth}
		\centering
		\includegraphics[width=0.25\linewidth]{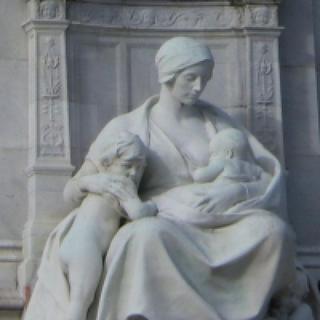}
		\vspace{-2mm}
		\caption*{\scriptsize input}
	\end{subfigure}
	\begin{subfigure}{0.24\linewidth}
		\centering
		\includegraphics[width=\linewidth]{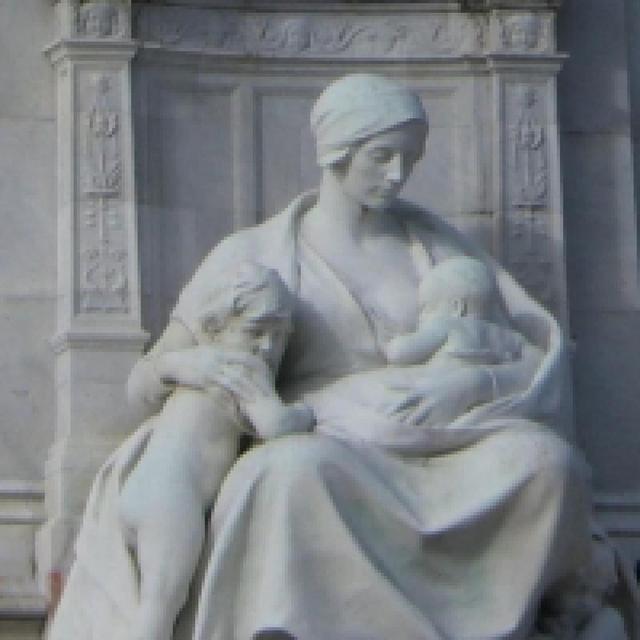}
		\vspace{-6mm}
		\caption*{\scriptsize bicubic interpolation}
	\end{subfigure}
	\begin{subfigure}{0.24\linewidth}
		\centering
		\includegraphics[width=\linewidth]{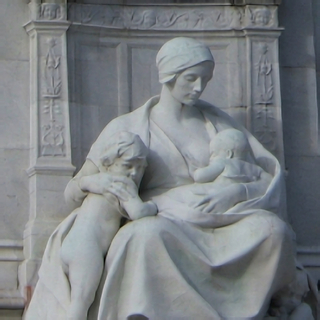}
		\vspace{-6mm}
		\caption*{\scriptsize \cite{FreFat10}}
	\end{subfigure}
	\begin{subfigure}{0.24\linewidth}
		\centering
		\includegraphics[width=\linewidth]{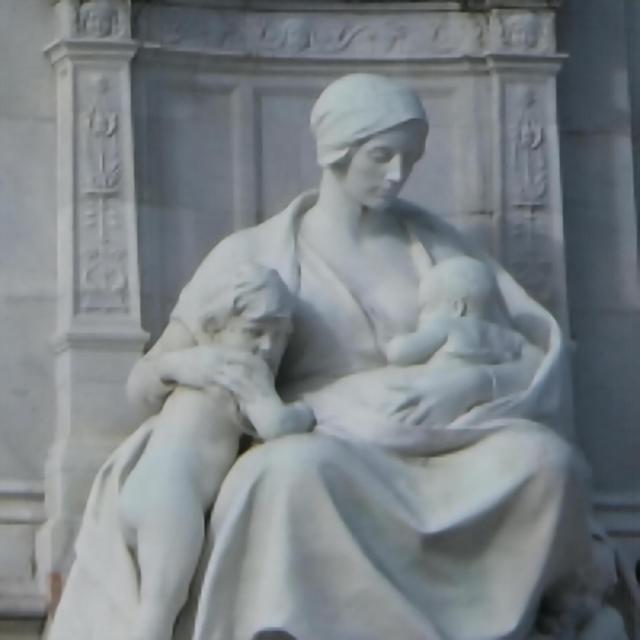}
		\vspace{-6mm}
		\caption*{\scriptsize proposed}
	\end{subfigure}
	\\
	\vspace{3mm}
	\def\idx{4}
	\begin{subfigure}{0.24\linewidth}
		\centering
		\includegraphics[width=0.125\linewidth]{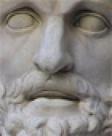}
		\vspace{-2mm}
		\caption*{\scriptsize input}
	\end{subfigure}
	\begin{subfigure}{0.24\linewidth}
		\centering
		\includegraphics[width=\linewidth]{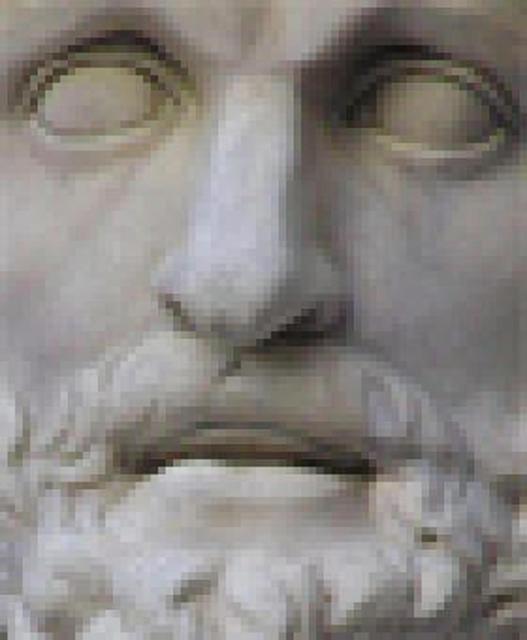}
		\vspace{-6mm}
		\caption*{\scriptsize bicubic interpolation}
	\end{subfigure}
	\begin{subfigure}{0.24\linewidth}
		\centering
		\includegraphics[width=\linewidth]{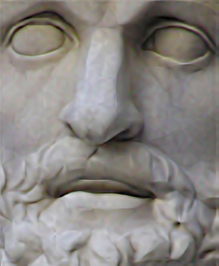}
		\vspace{-6mm}
		\caption*{\scriptsize \cite{FreFat10}}
	\end{subfigure}
	\begin{subfigure}{0.24\linewidth}
		\centering
		\includegraphics[width=\linewidth]{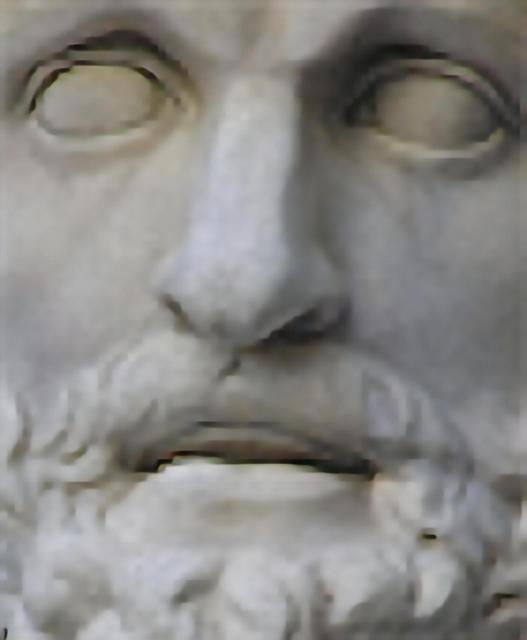}
		\vspace{-6mm}
		\caption*{\scriptsize proposed}
	\end{subfigure}
	
	\caption{Results of $4\times$ (on the first three rows) and $8\times$ (on the last row) super-resolution of Freeman and Fattal~\cite{FreFat10} (on the third column) and the proposed method (on the last column). All the input images are from~\cite{FreFat10}. Note that all the images, except for the one in the first row, do not have ground truth.   For the proposed method, we use the projection network trained on ImageNet dataset and set $\rho = 1.0$. }
	\label{figure: superres}
\end{figure*}

\myparagraph{Comparison to hand-designed signal priors.}   
We  compare the proposed framework with the traditional signal prior using $\ell_1$-norm of wavelet coefficients. 
We tune the weight of the $\ell_1$ prior, $\lambda$, based on the dataset. 
For image denoising, we compare with the state-of-the-art algorithm, BM3D~\cite{dabov2009bm3d}.
We add Gaussian random noise with different standard deviation $\sigma$ (out of $255$) to the test images, which were taken by the author with a cell phone camera.   
The value of $\sigma$ of each image is provided to BM3D.  
For the proposed method, we let $A = I$, the identity matrix, and set $\rho = \frac{3}{255} \sigma$. 
We use the same projection network learned from ImageNet dataset and apply it to $64 \times 64$ patches.
As shown in Figure~\ref{figure: denoise}, when $\sigma$ is larger than $40$, the proposed method consistently outperform BM3D.
For image super-resolution, we compare with the work of Freeman and Fattal~\cite{FreFat10}.
We perform $4\times$- and $8\times$-super resolution on the images from the result website of~\cite{FreFat10}.
The super-resolution results are shown in Figure~\ref{figure: superres}.


\def\dataset{celeb}
\def\denoise{denoise_ratio0.50_std0.10}
\def\denoiseours{ours_alpha0.300000}
\def\denoisel{l1_lambdal10.050000_alpha0.300000}
\def\cs{cs_ratio0.10_std0.00}
\def\csours{ours_alpha0.300000}
\def\csl{l1_lambdal10.050000_alpha0.300000}
\def\inpaintc{inpaintcenter_bs19_std0.00}
\def\inpaintcours{ours_alpha0.250000}
\def\inpaintcl{l1_lambdal10.050000_alpha0.300000}
\def\inpaint{inpaint_bs6_tb10_std0.00}
\def\inpaintours{ours_alpha0.300000}
\def\inpaintl{l1_lambdal10.050000_alpha0.300000}
\def\superres{superres_ratio0.50_std0.00}
\def\superresours{ours_alpha1.000000}
\def\superresl{l1_lambdal10.050000_alpha0.300000}
\def\superresquad{superres_ratio0.25_std0.00}
\def\superresquadours{ours_alpha1.000000}
\def\superresquadl{l1_lambdal10.050000_alpha0.300000}
\def\special{speical}

\begin{figure*}[thhhh]
	\centering
	\def\idxone{1630950}
	\def\idxtwo{46469}
	\begin{subfigure}{\linewidth}
		\centering
		\begin{subfigure}[t]{\thirteenimgwidth}
			\centering
			\caption*{\Cblankcol}
		\end{subfigure}	
		\begin{subfigure}[t]{\thirteenimgwidth}
			\centering
			\caption*{\Ccscol}
		\end{subfigure}	
		\begin{subfigure}[t]{\thirteenimgwidth}
			\centering
			\caption*{\Cdenoisecol}
		\end{subfigure}	
		\begin{subfigure}[t]{\thirteenimgwidth}
			\centering
			\caption*{\Cblockcol}
		\end{subfigure}	
		\begin{subfigure}[t]{\thirteenimgwidth}
			\centering
			\caption*{\Cinpaintcol}
		\end{subfigure}	
		\begin{subfigure}[t]{\thirteenimgwidth}
			\centering
			\caption*{\Csuperrescola}
		\end{subfigure}	
		\begin{subfigure}[t]{\thirteenimgwidth}
			\centering
			\caption*{\Csuperrescolb}
		\end{subfigure}	
		\hspace{0.5mm}
		\begin{subfigure}[t]{\thirteenimgwidth}
			\centering
			\caption*{\Ccscol}
		\end{subfigure}	
		\begin{subfigure}[t]{\thirteenimgwidth}
			\centering
			\caption*{\Cdenoisecol}
		\end{subfigure}	
		\begin{subfigure}[t]{\thirteenimgwidth}
			\centering
			\caption*{\Cblockcol}
		\end{subfigure}	
		\begin{subfigure}[t]{\thirteenimgwidth}
			\centering
			\caption*{\Cinpaintcol}
		\end{subfigure}	
		\begin{subfigure}[t]{\thirteenimgwidth}
			\centering
			\caption*{\Csuperrescola}
		\end{subfigure}	
		\begin{subfigure}[t]{\thirteenimgwidth}
			\centering
			\caption*{\Csuperrescolb}
		\end{subfigure}	
	\end{subfigure}
	\\
	\vspace{\thirteenimgrowmargin}
	\begin{subfigure}{\linewidth}
		\centering
		\begin{subfigure}[t]{\thirteenimgwidth}
			\centering
			\vspace{\thirteenimgrowcaptionheight}
			\caption*{\groundtruthrowcaption}
		\end{subfigure}	
		\begin{subfigure}[t]{\thirteenimgwidth}
			\centering
			\includegraphics[width=\textwidth]{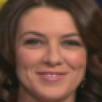}
		\end{subfigure}	
		\begin{subfigure}[t]{\thirteenimgwidth}
			\centering
			\includegraphics[width=\textwidth]{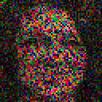}
		\end{subfigure}	
		\begin{subfigure}[t]{\thirteenimgwidth}
			\centering
			\includegraphics[width=\textwidth]{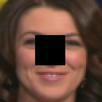}
		\end{subfigure}	
		\begin{subfigure}[t]{\thirteenimgwidth}
			\centering
			\includegraphics[width=\textwidth]{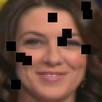}
		\end{subfigure}	
		\begin{subfigure}[t]{\thirteenimgwidth}
			\centering
			\vspace{-0.75\textwidth}
			\includegraphics[width=0.5\textwidth]{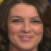}
		\end{subfigure}	
		\begin{subfigure}[t]{\thirteenimgwidth}
			\centering
			\vspace{-0.625\textwidth}
			\includegraphics[width=0.25\textwidth]{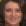}
		\end{subfigure}	
		\hspace{0.5mm}
		\begin{subfigure}[t]{\thirteenimgwidth}
			\centering
			\includegraphics[width=\textwidth]{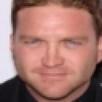}
		\end{subfigure}	
		\begin{subfigure}[t]{\thirteenimgwidth}
			\centering
			\includegraphics[width=\textwidth]{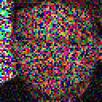}
		\end{subfigure}	
		\begin{subfigure}[t]{\thirteenimgwidth}
			\centering
			\includegraphics[width=\textwidth]{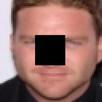}
		\end{subfigure}	
		\begin{subfigure}[t]{\thirteenimgwidth}
			\centering
			\includegraphics[width=\textwidth]{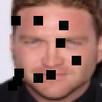}
		\end{subfigure}	
		\begin{subfigure}[t]{\thirteenimgwidth}
			\centering
			\vspace{-0.75\textwidth}
			\includegraphics[width=0.5\textwidth]{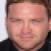}
		\end{subfigure}	
		\begin{subfigure}[t]{\thirteenimgwidth}
			\centering
			\vspace{-0.625\textwidth}
			\includegraphics[width=0.25\textwidth]{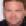}
		\end{subfigure}	
	\end{subfigure}
	\\
	\begin{subfigure}{\linewidth}
		\centering
		\begin{subfigure}[t]{\thirteenimgwidth}
			\centering
			\vspace{\thirteenimgrowcaptionheight}
			\caption*{\oursrowcaption}
		\end{subfigure}	
		\begin{subfigure}[t]{\thirteenimgwidth}
			\centering
			\includegraphics[width=\textwidth]{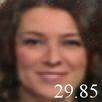}
		\end{subfigure}	
		\begin{subfigure}[t]{\thirteenimgwidth}
			\centering
			\includegraphics[width=\textwidth]{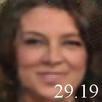}
		\end{subfigure}	
		\begin{subfigure}[t]{\thirteenimgwidth}
			\centering
			\includegraphics[width=\textwidth]{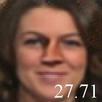}
		\end{subfigure}	
		\begin{subfigure}[t]{\thirteenimgwidth}
			\centering
			\includegraphics[width=\textwidth]{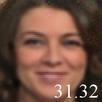}
		\end{subfigure}	
		\begin{subfigure}[t]{\thirteenimgwidth}
			\centering
			\includegraphics[width=\textwidth]{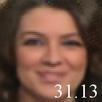}
		\end{subfigure}	
		\begin{subfigure}[t]{\thirteenimgwidth}
			\centering
			\includegraphics[width=\textwidth]{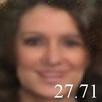}
		\end{subfigure}	
		\hspace{0.5mm}
		\begin{subfigure}[t]{\thirteenimgwidth}
			\centering
			\includegraphics[width=\textwidth]{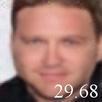}
		\end{subfigure}	
		\begin{subfigure}[t]{\thirteenimgwidth}
			\centering
			\includegraphics[width=\textwidth]{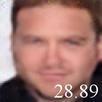}
		\end{subfigure}	
		\begin{subfigure}[t]{\thirteenimgwidth}
			\centering
			\includegraphics[width=\textwidth]{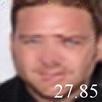}
		\end{subfigure}	
		\begin{subfigure}[t]{\thirteenimgwidth}
			\centering
			\includegraphics[width=\textwidth]{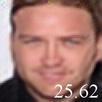}
		\end{subfigure}	
		\begin{subfigure}[t]{\thirteenimgwidth}
			\centering
			\includegraphics[width=\textwidth]{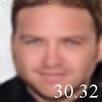}
		\end{subfigure}	
		\begin{subfigure}[t]{\thirteenimgwidth}
			\centering
			\includegraphics[width=\textwidth]{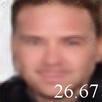}
		\end{subfigure}	
	\end{subfigure}
	\begin{subfigure}{\linewidth}
		\centering
		\begin{subfigure}[t]{\thirteenimgwidth}
			\centering
			\vspace{\thirteenimgrowcaptionheight}
			\caption*{\lrowcaption}
		\end{subfigure}	
		\begin{subfigure}[t]{\thirteenimgwidth}
			\centering
			\includegraphics[width=\textwidth]{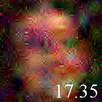}
		\end{subfigure}	
		\begin{subfigure}[t]{\thirteenimgwidth}
			\centering
			\includegraphics[width=\textwidth]{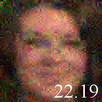}
		\end{subfigure}	
		\begin{subfigure}[t]{\thirteenimgwidth}
			\centering
			\includegraphics[width=\textwidth]{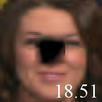}
		\end{subfigure}	
		\begin{subfigure}[t]{\thirteenimgwidth}
			\centering
			\includegraphics[width=\textwidth]{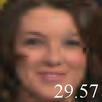}
		\end{subfigure}	
		\begin{subfigure}[t]{\thirteenimgwidth}
			\centering
			\includegraphics[width=\textwidth]{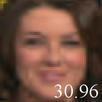}
		\end{subfigure}	
		\begin{subfigure}[t]{\thirteenimgwidth}
			\centering
			\includegraphics[width=\textwidth]{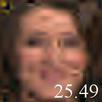}
		\end{subfigure}	
		\hspace{0.5mm}
		\begin{subfigure}[t]{\thirteenimgwidth}
			\centering
			\includegraphics[width=\textwidth]{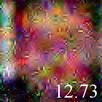}
		\end{subfigure}	
		\begin{subfigure}[t]{\thirteenimgwidth}
			\centering
			\includegraphics[width=\textwidth]{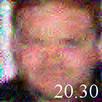}
		\end{subfigure}	
		\begin{subfigure}[t]{\thirteenimgwidth}
			\centering
			\includegraphics[width=\textwidth]{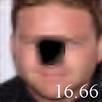}
		\end{subfigure}	
		\begin{subfigure}[t]{\thirteenimgwidth}
			\centering
			\includegraphics[width=\textwidth]{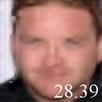}
		\end{subfigure}	
		\begin{subfigure}[t]{\thirteenimgwidth}
			\centering
			\includegraphics[width=\textwidth]{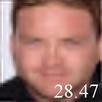}
		\end{subfigure}	
		\begin{subfigure}[t]{\thirteenimgwidth}
			\centering
			\includegraphics[width=\textwidth]{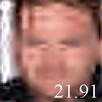}
		\end{subfigure}	
	\end{subfigure}
	\begin{subfigure}{\linewidth}
		\centering
		\begin{subfigure}[t]{\thirteenimgwidth}
			\centering
			\vspace{\thirteenimgrowcaptionheight}
			\caption*{\srowcaption}
		\end{subfigure}	
		\begin{subfigure}[t]{\thirteenimgwidth}
			\centering
			\includegraphics[width=\textwidth]{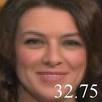}
		\end{subfigure}	
		\begin{subfigure}[t]{\thirteenimgwidth}
			\centering
			\includegraphics[width=\textwidth]{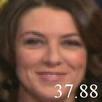}
		\end{subfigure}	
		\begin{subfigure}[t]{\thirteenimgwidth}
			\centering
			\includegraphics[width=\textwidth]{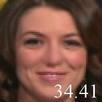}
		\end{subfigure}	
		\begin{subfigure}[t]{\thirteenimgwidth}
			\centering
			\includegraphics[width=\textwidth]{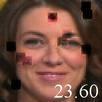}
		\end{subfigure}	
		\begin{subfigure}[t]{\thirteenimgwidth}
			\centering
			\includegraphics[width=\textwidth]{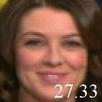}
		\end{subfigure}	
		\begin{subfigure}[t]{\thirteenimgwidth}
			\centering
			\includegraphics[width=\textwidth]{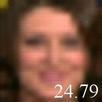}
		\end{subfigure}	
		\hspace{0.5mm}
		\begin{subfigure}[t]{\thirteenimgwidth}
			\centering
			\includegraphics[width=\textwidth]{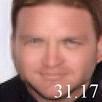}
		\end{subfigure}	
		\begin{subfigure}[t]{\thirteenimgwidth}
			\centering
			\includegraphics[width=\textwidth]{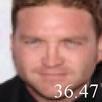}
		\end{subfigure}	
		\begin{subfigure}[t]{\thirteenimgwidth}
			\centering
			\includegraphics[width=\textwidth]{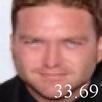}
		\end{subfigure}	
		\begin{subfigure}[t]{\thirteenimgwidth}
			\centering
			\includegraphics[width=\textwidth]{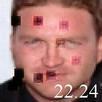}
		\end{subfigure}	
		\begin{subfigure}[t]{\thirteenimgwidth}
			\centering
			\includegraphics[width=\textwidth]{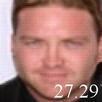}
		\end{subfigure}	
		\begin{subfigure}[t]{\thirteenimgwidth}
			\centering
			\includegraphics[width=\textwidth]{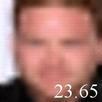}
		\end{subfigure}	
	\end{subfigure}
	\\
	\vspace{-1.5mm}
	\vspace{-2mm}
	\caption{Results on MS-Celeb-1M dataset. The PSNR values are shown in the lower-right corner of each image.  For compressive sensing, we test on  $\frac{m}{d} = 0.1$.  For  pixelwise inpainting, we drop $50\%$ of the pixels and add Gaussian noise with $\sigma=0.1$.  We use $\rho=1.0$ on both super resolution tasks.}
	\label{figure: celeb}
	\vspace{-2mm}
\end{figure*}

\def\dataset{imgnet}
\def\denoise{denoise_ratio0.50_std0.10}
\def\denoiseours{ours_alpha0.300000}
\def\denoisel{l1_lambdal10.100000_alpha0.300000}
\def\cs{cs_ratio0.10_std0.00}
\def\csours{ours_alpha0.300000}
\def\csl{l1_lambdal10.030000_alpha0.300000}
\def\inpaint{inpaint_bs6_tb10_std0.00}
\def\inpaintours{ours_alpha0.050000}
\def\inpaintl{l1_lambdal10.030000_alpha0.300000}
\def\inpaintc{inpaintcenter_bs19_std0.00}
\def\inpaintcours{ours_alpha0.300000}
\def\inpaintcl{l1_lambdal10.030000_alpha0.300000}
\def\superres{superres_ratio0.50_std0.00}
\def\superresours{ours_alpha0.500000}
\def\superresl{l1_lambdal10.030000_alpha0.500000}
\def\special{speical}

\begin{figure*}[t]
	\centering
	\def\idxone{905}
	\def\idxtwo{230}
	\def\idxthree{603}
	\def\idxfour{9783}
	\def\idxfive{12345}
	\def\idxsix{15213}
	\begin{subfigure}{\linewidth}
		\centering
		\begin{subfigure}[t]{\sixteenimgwidth}
			\centering
			\caption*{\Cblankcol}
		\end{subfigure}	
		\begin{subfigure}[t]{\sixteenimgwidth}
			\centering
			\caption*{\Ccscol}
		\end{subfigure}	
		\begin{subfigure}[t]{\sixteenimgwidth}
			\centering
			\caption*{\Cdenoisecol}
		\end{subfigure}	
		\begin{subfigure}[t]{\sixteenimgwidth}
			\centering
			\caption*{\Cinpaintcol}
		\end{subfigure}	
		\begin{subfigure}[t]{\sixteenimgwidth}
			\centering
			\caption*{\Cblockcol}
		\end{subfigure}	
		\begin{subfigure}[t]{\sixteenimgwidth}
			\centering
			\caption*{\Csuperrescola}
		\end{subfigure}	
		\hspace{0.5mm}
		\begin{subfigure}[t]{\sixteenimgwidth}
			\centering
			\caption*{\Ccscol}
		\end{subfigure}	
		\begin{subfigure}[t]{\sixteenimgwidth}
			\centering
			\caption*{\Cdenoisecol}
		\end{subfigure}	
		\begin{subfigure}[t]{\sixteenimgwidth}
			\centering
			\caption*{\Cinpaintcol}
		\end{subfigure}	
		\begin{subfigure}[t]{\sixteenimgwidth}
			\centering
			\caption*{\Cblockcol}
		\end{subfigure}	
		\begin{subfigure}[t]{\sixteenimgwidth}
			\centering
			\caption*{\Csuperrescola}
		\end{subfigure}	
		\hspace{0.5mm}
		\begin{subfigure}[t]{\sixteenimgwidth}
			\centering
			\caption*{\Ccscol}
		\end{subfigure}	
		\begin{subfigure}[t]{\sixteenimgwidth}
			\centering
			\caption*{\Cdenoisecol}
		\end{subfigure}	
		\begin{subfigure}[t]{\sixteenimgwidth}
			\centering
			\caption*{\Cinpaintcol}
		\end{subfigure}	
		\begin{subfigure}[t]{\sixteenimgwidth}
			\centering
			\caption*{\Cblockcol}
		\end{subfigure}	
		\begin{subfigure}[t]{\sixteenimgwidth}
			\centering
			\caption*{\Csuperrescola}
		\end{subfigure}	
	\end{subfigure}
	\\
	\vspace{\sixteenimgrowmargin}
	\begin{subfigure}{\linewidth}
		\centering
		\begin{subfigure}[t]{\sixteenimgwidth}
			\centering
			\vspace{\sixteenimgrowcaptionheight}
			\caption*{\groundtruthrowcaption}
		\end{subfigure}	
		\begin{subfigure}[t]{\sixteenimgwidth}
			\centering
			\includegraphics[width=\textwidth]{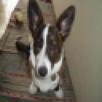}
		\end{subfigure}	
		\begin{subfigure}[t]{\sixteenimgwidth}
			\centering
			\includegraphics[width=\textwidth]{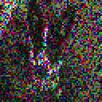}
		\end{subfigure}	
		\begin{subfigure}[t]{\sixteenimgwidth}
			\centering
			\includegraphics[width=\textwidth]{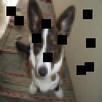}
		\end{subfigure}	
		\begin{subfigure}[t]{\sixteenimgwidth}
			\centering
			\includegraphics[width=\textwidth]{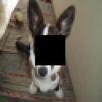}
		\end{subfigure}
		\begin{subfigure}[t]{\sixteenimgwidth}
			\centering
			\vspace{-0.75\textwidth}
			\includegraphics[width=0.5\textwidth]{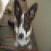}
		\end{subfigure}	
		\hspace{0.5mm}
		\begin{subfigure}[t]{\sixteenimgwidth}
			\centering
			\includegraphics[width=\textwidth]{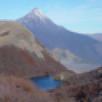}
		\end{subfigure}	
		\begin{subfigure}[t]{\sixteenimgwidth}
			\centering
			\includegraphics[width=\textwidth]{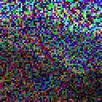}
		\end{subfigure}	
		\begin{subfigure}[t]{\sixteenimgwidth}
			\centering
			\includegraphics[width=\textwidth]{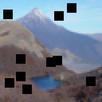}
		\end{subfigure}	
		\begin{subfigure}[t]{\sixteenimgwidth}
			\centering
			\includegraphics[width=\textwidth]{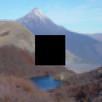}
		\end{subfigure}
		\begin{subfigure}[t]{\sixteenimgwidth}
			\centering
			\vspace{-0.75\textwidth}
			\includegraphics[width=0.5\textwidth]{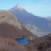}
		\end{subfigure}	
		\hspace{0.5mm}
		\begin{subfigure}[t]{\sixteenimgwidth}
			\centering
			\includegraphics[width=\textwidth]{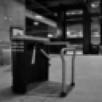}
		\end{subfigure}	
		\begin{subfigure}[t]{\sixteenimgwidth}
			\centering
			\includegraphics[width=\textwidth]{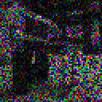}
		\end{subfigure}	
		\begin{subfigure}[t]{\sixteenimgwidth}
			\centering
			\includegraphics[width=\textwidth]{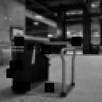}
		\end{subfigure}	
		\begin{subfigure}[t]{\sixteenimgwidth}
			\centering
			\includegraphics[width=\textwidth]{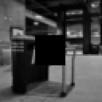}
		\end{subfigure}	
		\begin{subfigure}[t]{\sixteenimgwidth}
			\centering
			\vspace{-0.75\textwidth}
			\includegraphics[width=0.5\textwidth]{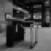}
		\end{subfigure}	
	\end{subfigure}	
	\\
	\begin{subfigure}{\linewidth}
		\centering
		\begin{subfigure}[t]{\sixteenimgwidth}
			\centering
			\vspace{\sixteenimgrowcaptionheight}
			\vspace{3mm}
			\caption*{\oursrowcaption}
		\end{subfigure}	
		\begin{subfigure}[t]{\sixteenimgwidth}
			\centering
			\includegraphics[width=\textwidth]{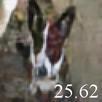}
		\end{subfigure}	
		\begin{subfigure}[t]{\sixteenimgwidth}
			\centering
			\includegraphics[width=\textwidth]{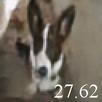}
		\end{subfigure}	
		\begin{subfigure}[t]{\sixteenimgwidth}
			\centering
			\includegraphics[width=\textwidth]{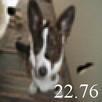}
		\end{subfigure}	
		\begin{subfigure}[t]{\sixteenimgwidth}
			\centering
			\includegraphics[width=\textwidth]{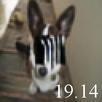}
		\end{subfigure}	
		\begin{subfigure}[t]{\sixteenimgwidth}
			\centering
			\includegraphics[width=\textwidth]{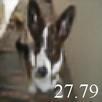}
		\end{subfigure}	
		\hspace{0.5mm}
		\begin{subfigure}[t]{\sixteenimgwidth}
			\centering
			\includegraphics[width=\textwidth]{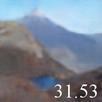}
		\end{subfigure}	
		\begin{subfigure}[t]{\sixteenimgwidth}
			\centering
			\includegraphics[width=\textwidth]{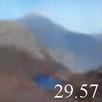}
		\end{subfigure}	
		\begin{subfigure}[t]{\sixteenimgwidth}
			\centering
			\includegraphics[width=\textwidth]{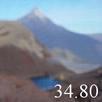}
		\end{subfigure}	
		\begin{subfigure}[t]{\sixteenimgwidth}
			\centering
			\includegraphics[width=\textwidth]{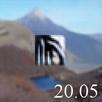}
		\end{subfigure}	
		\begin{subfigure}[t]{\sixteenimgwidth}
			\centering
			\includegraphics[width=\textwidth]{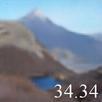}
		\end{subfigure}	
		\hspace{0.5mm}
		\begin{subfigure}[t]{\sixteenimgwidth}
			\centering
			\includegraphics[width=\textwidth]{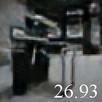}
		\end{subfigure}	
		\begin{subfigure}[t]{\sixteenimgwidth}
			\centering
			\includegraphics[width=\textwidth]{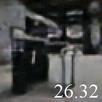}
		\end{subfigure}	
		\begin{subfigure}[t]{\sixteenimgwidth}
			\centering
			\includegraphics[width=\textwidth]{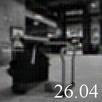}
		\end{subfigure}	
		\begin{subfigure}[t]{\sixteenimgwidth}
			\centering
			\includegraphics[width=\textwidth]{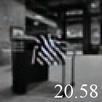}
		\end{subfigure}	
		\begin{subfigure}[t]{\sixteenimgwidth}
			\centering
			\includegraphics[width=\textwidth]{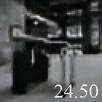}
		\end{subfigure}	
	\end{subfigure}
	\begin{subfigure}{\linewidth}
		\centering
		\begin{subfigure}[t]{\sixteenimgwidth}
			\centering
			\vspace{\sixteenimgrowcaptionheight}
			\vspace{3mm}
			\caption*{\lrowcaption}
		\end{subfigure}	
		\begin{subfigure}[t]{\sixteenimgwidth}
			\centering
			\includegraphics[width=\textwidth]{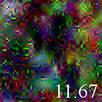}
		\end{subfigure}	
		\begin{subfigure}[t]{\sixteenimgwidth}
			\centering
			\includegraphics[width=\textwidth]{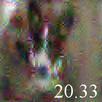}
		\end{subfigure}	
		\begin{subfigure}[t]{\sixteenimgwidth}
			\centering
			\includegraphics[width=\textwidth]{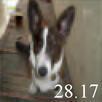}
		\end{subfigure}	
		\begin{subfigure}[t]{\sixteenimgwidth}
			\centering
			\includegraphics[width=\textwidth]{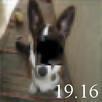}
		\end{subfigure}	
		\begin{subfigure}[t]{\sixteenimgwidth}
			\centering
			\includegraphics[width=\textwidth]{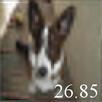}
		\end{subfigure}	
		\hspace{0.5mm}
		\begin{subfigure}[t]{\sixteenimgwidth}
			\centering
			\includegraphics[width=\textwidth]{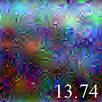}
		\end{subfigure}	
		\begin{subfigure}[t]{\sixteenimgwidth}
			\centering
			\includegraphics[width=\textwidth]{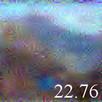}
		\end{subfigure}	
		\begin{subfigure}[t]{\sixteenimgwidth}
			\centering
			\includegraphics[width=\textwidth]{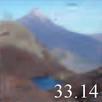}
		\end{subfigure}	
		\begin{subfigure}[t]{\sixteenimgwidth}
			\centering
			\includegraphics[width=\textwidth]{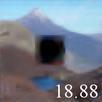}
		\end{subfigure}	
		\begin{subfigure}[t]{\sixteenimgwidth}
			\centering
			\includegraphics[width=\textwidth]{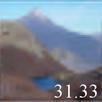}
		\end{subfigure}	
		\hspace{0.5mm}
		\begin{subfigure}[t]{\sixteenimgwidth}
			\centering
			\includegraphics[width=\textwidth]{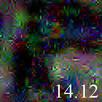}
		\end{subfigure}	
		\begin{subfigure}[t]{\sixteenimgwidth}
			\centering
			\includegraphics[width=\textwidth]{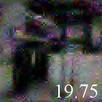}
		\end{subfigure}	
		\begin{subfigure}[t]{\sixteenimgwidth}
			\centering
			\includegraphics[width=\textwidth]{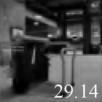}
		\end{subfigure}	
		\begin{subfigure}[t]{\sixteenimgwidth}
			\centering
			\includegraphics[width=\textwidth]{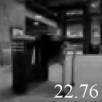}
		\end{subfigure}	
		\begin{subfigure}[t]{\sixteenimgwidth}
			\centering
			\includegraphics[width=\textwidth]{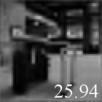}
		\end{subfigure}	
	\end{subfigure}
	\begin{subfigure}{\linewidth}
		\centering
		\begin{subfigure}[t]{\sixteenimgwidth}
			\centering
			\vspace{\sixteenimgrowcaptionheight}
			\caption*{\srowcaption}
		\end{subfigure}	
		\begin{subfigure}[t]{\sixteenimgwidth}
			\centering
			\includegraphics[width=\textwidth]{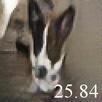}
		\end{subfigure}	
		\begin{subfigure}[t]{\sixteenimgwidth}
			\centering
			\includegraphics[width=\textwidth]{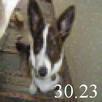}
		\end{subfigure}	
		\begin{subfigure}[t]{\sixteenimgwidth}
			\centering
			\includegraphics[width=\textwidth]{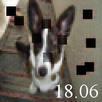}
		\end{subfigure}	
		\begin{subfigure}[t]{\sixteenimgwidth}
			\centering
			\includegraphics[width=\textwidth]{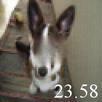}
		\end{subfigure}	
		\begin{subfigure}[t]{\sixteenimgwidth}
			\centering
			\includegraphics[width=\textwidth]{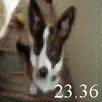}
		\end{subfigure}	
		\hspace{0.5mm}
		\begin{subfigure}[t]{\sixteenimgwidth}
			\centering
			\includegraphics[width=\textwidth]{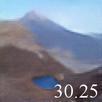}
		\end{subfigure}	
		\begin{subfigure}[t]{\sixteenimgwidth}
			\centering
			\includegraphics[width=\textwidth]{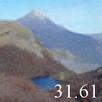}
		\end{subfigure}	
		\begin{subfigure}[t]{\sixteenimgwidth}
			\centering
			\includegraphics[width=\textwidth]{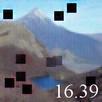}
		\end{subfigure}	
		\begin{subfigure}[t]{\sixteenimgwidth}
			\centering
			\includegraphics[width=\textwidth]{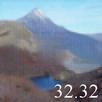}
		\end{subfigure}	
		\begin{subfigure}[t]{\sixteenimgwidth}
			\centering
			\includegraphics[width=\textwidth]{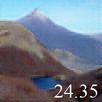}
		\end{subfigure}	
		\hspace{0.5mm}
		\begin{subfigure}[t]{\sixteenimgwidth}
			\centering
			\includegraphics[width=\textwidth]{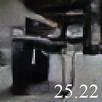}
		\end{subfigure}	
		\begin{subfigure}[t]{\sixteenimgwidth}
			\centering
			\includegraphics[width=\textwidth]{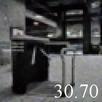}
		\end{subfigure}	
		\begin{subfigure}[t]{\sixteenimgwidth}
			\centering
			\includegraphics[width=\textwidth]{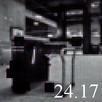}
		\end{subfigure}	
		\begin{subfigure}[t]{\sixteenimgwidth}
			\centering
			\includegraphics[width=\textwidth]{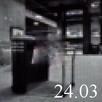}
		\end{subfigure}
		\begin{subfigure}[t]{\sixteenimgwidth}
			\centering
			\includegraphics[width=\textwidth]{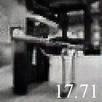}
		\end{subfigure}	
	\end{subfigure}
	\def\idxone{\idxfour}
	\def\idxtwo{\idxfive}
	\def\idxthree{\idxsix}
	\vspace{\sixteenimgrowmargin}
	\begin{subfigure}{\linewidth}
		\centering
		\begin{subfigure}[t]{\sixteenimgwidth}
			\centering
			\vspace{\sixteenimgrowcaptionheight}
			\caption*{\groundtruthrowcaption}
		\end{subfigure}	
		\begin{subfigure}[t]{\sixteenimgwidth}
			\centering
			\includegraphics[width=\textwidth]{figures/\dataset/\idxone/ori_img.jpg}
		\end{subfigure}	
		\begin{subfigure}[t]{\sixteenimgwidth}
			\centering
			\includegraphics[width=\textwidth]{figures/\dataset/\idxone/\denoise/y.jpg}
		\end{subfigure}	
		\begin{subfigure}[t]{\sixteenimgwidth}
			\centering
			\includegraphics[width=\textwidth]{figures/{\dataset}/{\idxone}/{\inpaint}/y.jpg}
		\end{subfigure}	
		\begin{subfigure}[t]{\sixteenimgwidth}
			\centering
			\includegraphics[width=\textwidth]{figures/{\dataset}/{\idxone}/{\inpaintc}/y.jpg}
		\end{subfigure}
		\begin{subfigure}[t]{\sixteenimgwidth}
			\centering
			\vspace{-0.75\textwidth}
			\includegraphics[width=0.5\textwidth]{figures/\dataset/\idxone/\superres/y.jpg}
		\end{subfigure}	
		\hspace{0.5mm}
		\begin{subfigure}[t]{\sixteenimgwidth}
			\centering
			\includegraphics[width=\textwidth]{figures/\dataset/\idxtwo/ori_img.jpg}
		\end{subfigure}	
		\begin{subfigure}[t]{\sixteenimgwidth}
			\centering
			\includegraphics[width=\textwidth]{figures/\dataset/\idxtwo/\denoise/y.jpg}
		\end{subfigure}	
		\begin{subfigure}[t]{\sixteenimgwidth}
			\centering
			\includegraphics[width=\textwidth]{figures/{\dataset}/{\idxtwo}/{\inpaint}/y.jpg}
		\end{subfigure}	
		\begin{subfigure}[t]{\sixteenimgwidth}
			\centering
			\includegraphics[width=\textwidth]{figures/{\dataset}/{\idxtwo}/{\inpaintc}/y.jpg}
		\end{subfigure}
		\begin{subfigure}[t]{\sixteenimgwidth}
			\centering
			\vspace{-0.75\textwidth}
			\includegraphics[width=0.5\textwidth]{figures/\dataset/\idxtwo/\superres/y.jpg}
		\end{subfigure}	
		\hspace{0.5mm}
		\begin{subfigure}[t]{\sixteenimgwidth}
			\centering
			\includegraphics[width=\textwidth]{figures/\dataset/\idxthree/ori_img.jpg}
		\end{subfigure}	
		\begin{subfigure}[t]{\sixteenimgwidth}
			\centering
			\includegraphics[width=\textwidth]{figures/\dataset/\idxthree/\denoise/y.jpg}
		\end{subfigure}	
		\begin{subfigure}[t]{\sixteenimgwidth}
			\centering
			\includegraphics[width=\textwidth]{figures/{\dataset}/{\idxthree}/{\inpaint}/y.jpg}
		\end{subfigure}	
		\begin{subfigure}[t]{\sixteenimgwidth}
			\centering
			\includegraphics[width=\textwidth]{figures/{\dataset}/{\idxthree}/{\inpaintc}/y.jpg}
		\end{subfigure}	
		\begin{subfigure}[t]{\sixteenimgwidth}
			\centering
			\vspace{-0.75\textwidth}
			\includegraphics[width=0.5\textwidth]{figures/\dataset/\idxthree/\superres/y.jpg}
		\end{subfigure}	
	\end{subfigure}	
	\\
	\begin{subfigure}{\linewidth}
		\centering
		\begin{subfigure}[t]{\sixteenimgwidth}
			\centering
			\vspace{\sixteenimgrowcaptionheight}
			\vspace{3mm}
			\caption*{\oursrowcaption}
		\end{subfigure}	
		\begin{subfigure}[t]{\sixteenimgwidth}
			\centering
			\includegraphics[width=\textwidth]{figures/\dataset/\idxone/\cs/\csours/z.jpg}
		\end{subfigure}	
		\begin{subfigure}[t]{\sixteenimgwidth}
			\centering
			\includegraphics[width=\textwidth]{figures/\dataset/\idxone/\denoise/\denoiseours/z.jpg}
		\end{subfigure}	
		\begin{subfigure}[t]{\sixteenimgwidth}
			\centering
			\includegraphics[width=\textwidth]{figures/{\dataset}/{\idxone}/{\inpaint}/\inpaintours/z.jpg}
		\end{subfigure}	
		\begin{subfigure}[t]{\sixteenimgwidth}
			\centering
			\includegraphics[width=\textwidth]{figures/{\dataset}/{\idxone}/{\inpaintc}/\inpaintcours/z.jpg}
		\end{subfigure}	
		\begin{subfigure}[t]{\sixteenimgwidth}
			\centering
			\includegraphics[width=\textwidth]{figures/\dataset/\idxone/\superres/\superresours/z.jpg}
		\end{subfigure}	
		\hspace{0.5mm}
		\begin{subfigure}[t]{\sixteenimgwidth}
			\centering
			\includegraphics[width=\textwidth]{figures/\dataset/\idxtwo/\cs/\csours/z.jpg}
		\end{subfigure}	
		\begin{subfigure}[t]{\sixteenimgwidth}
			\centering
			\includegraphics[width=\textwidth]{figures/\dataset/\idxtwo/\denoise/\denoiseours/z.jpg}
		\end{subfigure}	
		\begin{subfigure}[t]{\sixteenimgwidth}
			\centering
			\includegraphics[width=\textwidth]{figures/{\dataset}/{\idxtwo}/{\inpaint}/\inpaintours/z.jpg}
		\end{subfigure}	
		\begin{subfigure}[t]{\sixteenimgwidth}
			\centering
			\includegraphics[width=\textwidth]{figures/{\dataset}/{\idxtwo}/{\inpaintc}/\inpaintcours/z.jpg}
		\end{subfigure}	
		\begin{subfigure}[t]{\sixteenimgwidth}
			\centering
			\includegraphics[width=\textwidth]{figures/\dataset/\idxtwo/\superres/\superresours/z.jpg}
		\end{subfigure}	
		\hspace{0.5mm}
		\begin{subfigure}[t]{\sixteenimgwidth}
			\centering
			\includegraphics[width=\textwidth]{figures/\dataset/\idxthree/\cs/\csours/z.jpg}
		\end{subfigure}	
		\begin{subfigure}[t]{\sixteenimgwidth}
			\centering
			\includegraphics[width=\textwidth]{figures/\dataset/\idxthree/\denoise/\denoiseours/z.jpg}
		\end{subfigure}	
		\begin{subfigure}[t]{\sixteenimgwidth}
			\centering
			\includegraphics[width=\textwidth]{figures/{\dataset}/{\idxthree}/{\inpaint}/\inpaintours/z.jpg}
		\end{subfigure}	
		\begin{subfigure}[t]{\sixteenimgwidth}
			\centering
			\includegraphics[width=\textwidth]{figures/{\dataset}/{\idxthree}/{\inpaintc}/\inpaintcours/z.jpg}
		\end{subfigure}	
		\begin{subfigure}[t]{\sixteenimgwidth}
			\centering
			\includegraphics[width=\textwidth]{figures/\dataset/\idxthree/\superres/\superresours/z.jpg}
		\end{subfigure}	
	\end{subfigure}
	\begin{subfigure}{\linewidth}
		\centering
		\begin{subfigure}[t]{\sixteenimgwidth}
			\centering
			\vspace{\sixteenimgrowcaptionheight}
			\vspace{3mm}
			\caption*{\lrowcaption}
		\end{subfigure}	
		\begin{subfigure}[t]{\sixteenimgwidth}
			\centering
			\includegraphics[width=\textwidth]{figures/\dataset/\idxone/\cs/\csl/z.jpg}
		\end{subfigure}	
		\begin{subfigure}[t]{\sixteenimgwidth}
			\centering
			\includegraphics[width=\textwidth]{figures/\dataset/\idxone/\denoise/\denoisel/z.jpg}
		\end{subfigure}	
		\begin{subfigure}[t]{\sixteenimgwidth}
			\centering
			\includegraphics[width=\textwidth]{figures/{\dataset}/{\idxone}/{\inpaint}/\inpaintl/z.jpg}
		\end{subfigure}	
		\begin{subfigure}[t]{\sixteenimgwidth}
			\centering
			\includegraphics[width=\textwidth]{figures/{\dataset}/{\idxone}/{\inpaintc}/\inpaintcl/z.jpg}
		\end{subfigure}	
		\begin{subfigure}[t]{\sixteenimgwidth}
			\centering
			\includegraphics[width=\textwidth]{figures/\dataset/\idxone/\superres/\superresl/z.jpg}
		\end{subfigure}	
		\hspace{0.5mm}
		\begin{subfigure}[t]{\sixteenimgwidth}
			\centering
			\includegraphics[width=\textwidth]{figures/\dataset/\idxtwo/\cs/\csl/z.jpg}
		\end{subfigure}	
		\begin{subfigure}[t]{\sixteenimgwidth}
			\centering
			\includegraphics[width=\textwidth]{figures/\dataset/\idxtwo/\denoise/\denoisel/z.jpg}
		\end{subfigure}	
		\begin{subfigure}[t]{\sixteenimgwidth}
			\centering
			\includegraphics[width=\textwidth]{figures/{\dataset}/{\idxtwo}/{\inpaint}/\inpaintl/z.jpg}
		\end{subfigure}	
		\begin{subfigure}[t]{\sixteenimgwidth}
			\centering
			\includegraphics[width=\textwidth]{figures/{\dataset}/{\idxtwo}/{\inpaintc}/\inpaintcl/z.jpg}
		\end{subfigure}	
		\begin{subfigure}[t]{\sixteenimgwidth}
			\centering
			\includegraphics[width=\textwidth]{figures/\dataset/\idxtwo/\superres/\superresl/z.jpg}
		\end{subfigure}	
		\hspace{0.5mm}
		\begin{subfigure}[t]{\sixteenimgwidth}
			\centering
			\includegraphics[width=\textwidth]{figures/\dataset/\idxthree/\cs/\csl/z.jpg}
		\end{subfigure}	
		\begin{subfigure}[t]{\sixteenimgwidth}
			\centering
			\includegraphics[width=\textwidth]{figures/\dataset/\idxthree/\denoise/\denoisel/z.jpg}
		\end{subfigure}	
		\begin{subfigure}[t]{\sixteenimgwidth}
			\centering
			\includegraphics[width=\textwidth]{figures/{\dataset}/{\idxthree}/{\inpaint}/\inpaintl/z.jpg}
		\end{subfigure}	
		\begin{subfigure}[t]{\sixteenimgwidth}
			\centering
			\includegraphics[width=\textwidth]{figures/{\dataset}/{\idxthree}/{\inpaintc}/\inpaintcl/z.jpg}
		\end{subfigure}	
		\begin{subfigure}[t]{\sixteenimgwidth}
			\centering
			\includegraphics[width=\textwidth]{figures/\dataset/\idxthree/\superres/\superresl/z.jpg}
		\end{subfigure}	
	\end{subfigure}
	\begin{subfigure}{\linewidth}
		\centering
		\begin{subfigure}[t]{\sixteenimgwidth}
			\centering
			\vspace{\sixteenimgrowcaptionheight}
			\caption*{\srowcaption}
		\end{subfigure}	
		\begin{subfigure}[t]{\sixteenimgwidth}
			\centering
			\includegraphics[width=\textwidth]{figures/\dataset/\idxone/\cs/\special/z.jpg}
		\end{subfigure}	
		\begin{subfigure}[t]{\sixteenimgwidth}
			\centering
			\includegraphics[width=\textwidth]{figures/\dataset/\idxone/\denoise/\special/z.jpg}
		\end{subfigure}	
		\begin{subfigure}[t]{\sixteenimgwidth}
			\centering
			\includegraphics[width=\textwidth]{figures/{\dataset}/{\idxone}/{\inpaint}/\special/z.jpg}
		\end{subfigure}	
		\begin{subfigure}[t]{\sixteenimgwidth}
			\centering
			\includegraphics[width=\textwidth]{figures/{\dataset}/{\idxone}/{\inpaintc}/\special/z.jpg}
		\end{subfigure}	
		\begin{subfigure}[t]{\sixteenimgwidth}
			\centering
			\includegraphics[width=\textwidth]{figures/\dataset/\idxone/\superres/\special/z.jpg}
		\end{subfigure}	
		\hspace{0.5mm}
		\begin{subfigure}[t]{\sixteenimgwidth}
			\centering
			\includegraphics[width=\textwidth]{figures/\dataset/\idxtwo/\cs/\special/z.jpg}
		\end{subfigure}	
		\begin{subfigure}[t]{\sixteenimgwidth}
			\centering
			\includegraphics[width=\textwidth]{figures/\dataset/\idxtwo/\denoise/\special/z.jpg}
		\end{subfigure}	
		\begin{subfigure}[t]{\sixteenimgwidth}
			\centering
			\includegraphics[width=\textwidth]{figures/{\dataset}/{\idxtwo}/{\inpaint}/\special/z.jpg}
		\end{subfigure}	
		\begin{subfigure}[t]{\sixteenimgwidth}
			\centering
			\includegraphics[width=\textwidth]{figures/{\dataset}/{\idxtwo}/{\inpaintc}/\special/z.jpg}
		\end{subfigure}	
		\begin{subfigure}[t]{\sixteenimgwidth}
			\centering
			\includegraphics[width=\textwidth]{figures/\dataset/\idxtwo/\superres/\special/z.jpg}
		\end{subfigure}	
		\hspace{0.5mm}
		\begin{subfigure}[t]{\sixteenimgwidth}
			\centering
			\includegraphics[width=\textwidth]{figures/\dataset/\idxthree/\cs/\special/z.jpg}
		\end{subfigure}	
		\begin{subfigure}[t]{\sixteenimgwidth}
			\centering
			\includegraphics[width=\textwidth]{figures/\dataset/\idxthree/\denoise/\special/z.jpg}
		\end{subfigure}	
		\begin{subfigure}[t]{\sixteenimgwidth}
			\centering
			\includegraphics[width=\textwidth]{figures/{\dataset}/{\idxthree}/{\inpaint}/\special/z.jpg}
		\end{subfigure}	
		\begin{subfigure}[t]{\sixteenimgwidth}
			\centering
			\includegraphics[width=\textwidth]{figures/{\dataset}/{\idxthree}/{\inpaintc}/\special/z.jpg}
		\end{subfigure}
		\begin{subfigure}[t]{\sixteenimgwidth}
			\centering
			\includegraphics[width=\textwidth]{figures/\dataset/\idxthree/\superres/\special/z.jpg}
		\end{subfigure}	
	\end{subfigure}
	\vspace{-2mm}
	\caption{Results on ImageNet dataset. The PSNR values are shown in the lower-right corner of each image.  Compressive sensing uses $\frac{m}{d} = 0.1$.  For pixelwise inpainting, we drop $50\%$ of the pixels and add Gaussian noise with $\sigma=0.1$.  We use $\rho=0.05$ on scattered inpainting and $\rho=0.5$ on super resolution.  }
	\label{figure: imgnet}
\end{figure*}

\def\dataset{cs_results}

\begin{figure*}[t]
	\centering
	\def\idx{21222}
		\begin{subfigure}[t]{\twelveimgwidth}
			\centering
			\caption*{\scriptsize ground truth}
		\end{subfigure}	
		\hspace{0.5mm}
		\begin{subfigure}[t]{\twelveimgwidth}
			\centering
			\caption*{\scriptsize orignal result}
		\end{subfigure}	
		\begin{subfigure}[t]{\twelveimgwidth}
			\centering
			\caption*{\scriptsize resample $1\%$}
		\end{subfigure}	
		\begin{subfigure}[t]{\twelveimgwidth}
			\centering
			\caption*{\scriptsize resample $5\%$}
		\end{subfigure}	
		\begin{subfigure}[t]{\twelveimgwidth}
			\centering
			\caption*{\scriptsize resample $10\%$}
		\end{subfigure}	
		\begin{subfigure}[t]{\twelveimgwidth}
			\centering
			\caption*{\scriptsize resample $20\%$}
		\end{subfigure}	
		\begin{subfigure}[t]{\twelveimgwidth}
			\centering
			\caption*{\scriptsize noise $\sigma{=} 0.1$}
		\end{subfigure}	
		\begin{subfigure}[t]{\twelveimgwidth}
			\centering
			\caption*{\scriptsize noise $\sigma{=} 0.2$}
		\end{subfigure}	
		\begin{subfigure}[t]{\twelveimgwidth}
			\centering
			\caption*{\scriptsize noise $\sigma{=} 0.3$}
		\end{subfigure}	
		\begin{subfigure}[t]{\twelveimgwidth}
			\centering
			\caption*{\scriptsize noise $\sigma{=} 0.4$}
		\end{subfigure}	
		\begin{subfigure}[t]{\twelveimgwidth}
			\centering
			\caption*{\scriptsize noise $\sigma{=} 0.5$}
		\end{subfigure}	
	\\
		\vspace{\twelveimgrowmargin}
		\begin{subfigure}[t]{\twelveimgwidth}
			\includegraphics[width=\linewidth]{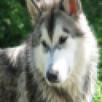}
		\end{subfigure}
		\hspace{0.5mm}
		\begin{subfigure}[t]{\twelveimgwidth}
			\includegraphics[width=\linewidth]{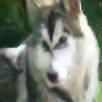}
			\centering
			\vspace{-6mm}
			\caption*{\scriptsize 24.45}
		\end{subfigure}
		\begin{subfigure}[t]{\twelveimgwidth}
			\includegraphics[width=\linewidth]{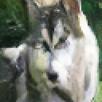}
			\centering
			\vspace{-6mm}
			\caption*{\scriptsize 22.48}
		\end{subfigure}
		\begin{subfigure}[t]{\twelveimgwidth}
			\includegraphics[width=\linewidth]{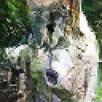}
			\centering
			\vspace{-6mm}
			\caption*{\scriptsize 17.95}
		\end{subfigure}
		\begin{subfigure}[t]{\twelveimgwidth}
			\includegraphics[width=\linewidth]{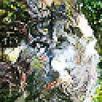}
			\centering
			\vspace{-6mm}
			\caption*{\scriptsize 14.48}
		\end{subfigure}
		\begin{subfigure}[t]{\twelveimgwidth}
			\includegraphics[width=\linewidth]{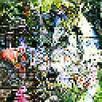}
			\centering
			\vspace{-6mm}
			\caption*{\scriptsize 11.51}
		\end{subfigure}
		\hspace{0.5mm}
		\begin{subfigure}[t]{\twelveimgwidth}
			\includegraphics[width=\linewidth]{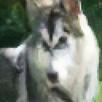}
			\centering
			\vspace{-6mm}
			\caption*{\scriptsize 23.67}
		\end{subfigure}
		\begin{subfigure}[t]{\twelveimgwidth}
			\includegraphics[width=\linewidth]{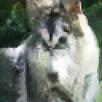}
			\centering
			\vspace{-6mm}
			\caption*{\scriptsize 21.72}
		\end{subfigure}
		\begin{subfigure}[t]{\twelveimgwidth}
			\includegraphics[width=\linewidth]{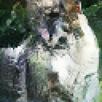}
			\centering
			\vspace{-6mm}
			\caption*{\scriptsize 19.26}
		\end{subfigure}
		\begin{subfigure}[t]{\twelveimgwidth}
			\includegraphics[width=\linewidth]{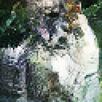}
			\centering
			\vspace{-6mm}
			\caption*{\scriptsize 17.10}
		\end{subfigure}
		\begin{subfigure}[t]{\twelveimgwidth}
			\includegraphics[width=\linewidth]{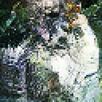}
			\centering
			\vspace{-6mm}
			\caption*{\scriptsize 15.47}
		\end{subfigure}
	\\
		\vspace{1.5mm}
		\begin{subfigure}[t]{\twelveimgwidth}
			\hspace{\linewidth}
		\end{subfigure}
	    \hspace{0.5mm}
		\begin{subfigure}[t]{\twelveimgwidth}
			\includegraphics[width=\linewidth]{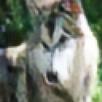}
			\centering
			\vspace{-6mm}
			\caption*{\scriptsize 24.14}
		\end{subfigure}
		\begin{subfigure}[t]{\twelveimgwidth}
			\includegraphics[width=\linewidth]{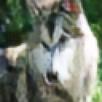}
			\centering
			\vspace{-6mm}
			\caption*{\scriptsize 24.17}
		\end{subfigure}
		\begin{subfigure}[t]{\twelveimgwidth}
			\includegraphics[width=\linewidth]{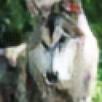}
			\centering
			\vspace{-6mm}
			\caption*{\scriptsize 24.47}
		\end{subfigure}
		\begin{subfigure}[t]{\twelveimgwidth}
			\includegraphics[width=\linewidth]{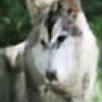}
			\centering
			\vspace{-6mm}
			\caption*{\scriptsize 23.18}
		\end{subfigure}
		\begin{subfigure}[t]{\twelveimgwidth}
			\includegraphics[width=\linewidth]{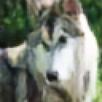}
			\centering
			\vspace{-6mm}
			\caption*{\scriptsize 24.66}
		\end{subfigure}
		\hspace{0.5mm}
		\begin{subfigure}[t]{\twelveimgwidth}
			\includegraphics[width=\linewidth]{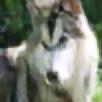}
			\centering
			\vspace{-6mm}
			\caption*{\scriptsize 24.44}
		\end{subfigure}
		\begin{subfigure}[t]{\twelveimgwidth}
			\includegraphics[width=\linewidth]{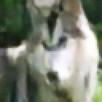}
			\centering
			\vspace{-6mm}
			\caption*{\scriptsize 23.49}
		\end{subfigure}
		\begin{subfigure}[t]{\twelveimgwidth}
			\includegraphics[width=\linewidth]{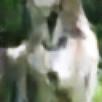}
			\centering
			\vspace{-6mm}
			\caption*{\scriptsize 22.37}
		\end{subfigure}
		\begin{subfigure}[t]{\twelveimgwidth}
			\includegraphics[width=\linewidth]{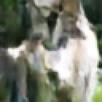}
			\centering
			\vspace{-6mm}
			\caption*{\scriptsize 20.39}
		\end{subfigure}
		\begin{subfigure}[t]{\twelveimgwidth}
			\includegraphics[width=\linewidth]{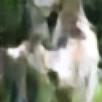}
			\centering
			\vspace{-6mm}
			\caption*{\scriptsize 20.50}
		\end{subfigure}
	\vspace{-3mm}
	\caption{\small Comparison on the robustness to the linear operator $A$ and noise on compressive sensing.  We start from the training measurement matrix $A$ of the specialized network and randomly resample some elements in $A$.  We also add Gaussian noise with different standard deviation $\sigma$ to the linear measurements $\y$ of original $A$.  The upper row shows the results of the specialized trained network, and the bottom one demonstrates results from the proposed framework.  We use $\rho=0.5$ for $\sigma=0.2$, $\rho=0.7$ for $\sigma=0.3$, $\rho=1.0$ for $\sigma=0.4$, $\rho=1.1$ for $\sigma=0.5$, and $\rho=0.3$ for all other cases. The PSNR values are shown in the below of each result.}
	\label{figure: cs compare}
\end{figure*}

\def\dataset{denoise_images_clean}
\def\imgwidth{0.24\linewidth}
\def\capwidth{-6mm}

\begin{figure*}[t]
	\centering
	\def\idxone{1456}
	\begin{subfigure}{\linewidth}
		\centering
		\begin{subfigure}[t]{\imgwidth}
			\centering
			\includegraphics[width=\textwidth]{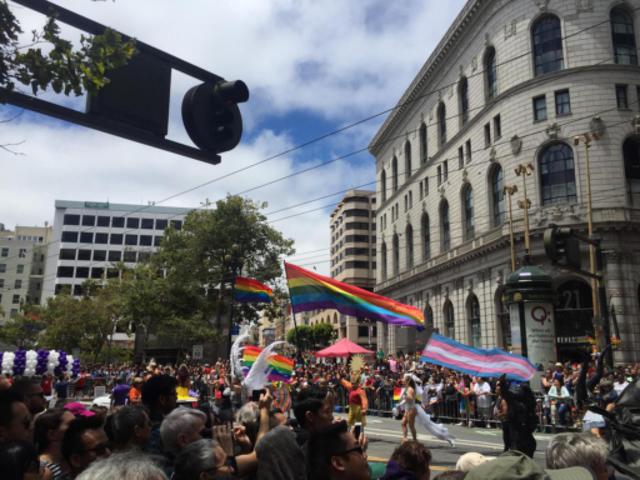}
			\vspace{\capwidth}
			\caption*{\scriptsize ground truth}
		\end{subfigure}	
		\def\imgsigma{40}
		\def\ouralpha{0.470588}
		\def\bm3dpsnr{26.56}
		\def\ourspsnr{27.47}
		\begin{subfigure}[t]{\imgwidth}
			\centering
			\includegraphics[width=\textwidth]{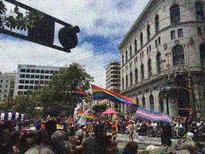}
			\vspace{\capwidth}
			\caption*{\scriptsize input ($\sigma=\imgsigma$)}
		\end{subfigure}	
		\begin{subfigure}[t]{\imgwidth}
			\centering
			\includegraphics[width=\textwidth]{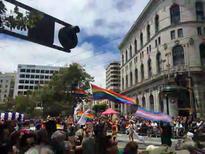}
			\vspace{\capwidth}
			\caption*{\scriptsize BM3D (PSNR=$\bm3dpsnr$)}
		\end{subfigure}
		\begin{subfigure}[t]{\imgwidth}
			\centering
			\includegraphics[width=\textwidth]{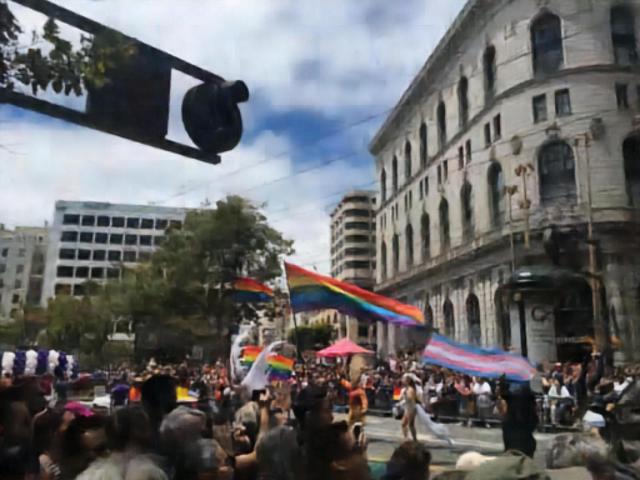}
			\vspace{\capwidth}
			\caption*{\scriptsize proposed (PSNR=$\ourspsnr$)}
		\end{subfigure}
		\\
		\vspace{1mm}
		\begin{subfigure}[t]{\imgwidth}
			\centering
			\includegraphics[width=\textwidth]{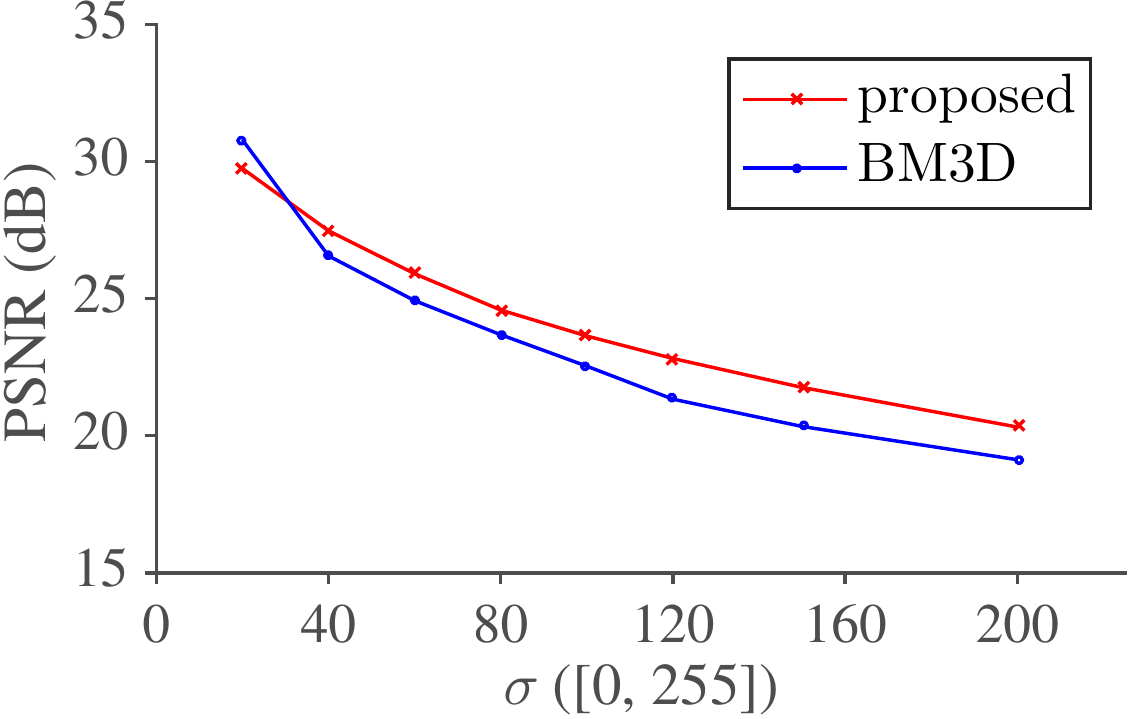}
			\vspace{\capwidth}
			\caption*{\scriptsize PSNR \vs $\sigma$}
		\end{subfigure}	
		\def\imgsigma{100}
		\def\ouralpha{1.176471}
		\def\bm3dpsnr{22.54}
		\def\ourspsnr{23.64}
		\begin{subfigure}[t]{\imgwidth}
			\centering
			\includegraphics[width=\textwidth]{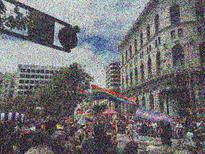}
			\vspace{\capwidth}
			\caption*{\scriptsize input ($\sigma=\imgsigma$)}
		\end{subfigure}	
		\begin{subfigure}[t]{\imgwidth}
			\centering
			\includegraphics[width=\textwidth]{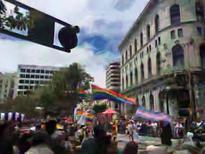}
			\vspace{\capwidth}
			\caption*{\scriptsize BM3D (PSNR=$\bm3dpsnr$)}
		\end{subfigure}
		\begin{subfigure}[t]{\imgwidth}
			\centering
			\includegraphics[width=\textwidth]{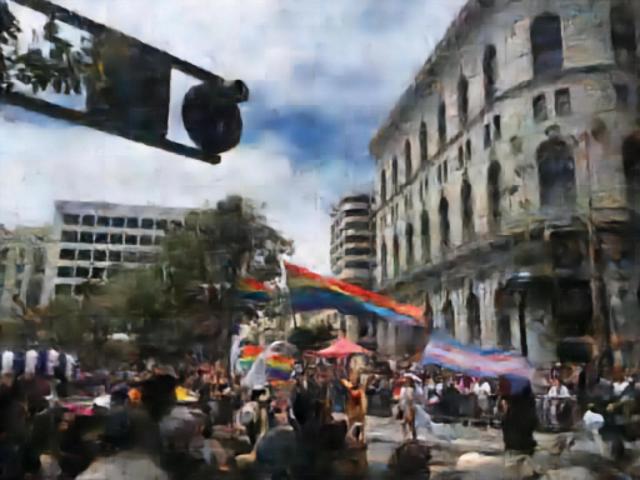}
			\vspace{\capwidth}
			\caption*{\scriptsize proposed (PSNR=$\ourspsnr$)}
		\end{subfigure}
		\\
		\vspace{1mm}
		\begin{subfigure}[t]{\imgwidth}
			\centering
			\hspace{\textwidth}
		\end{subfigure}	
		\def\imgsigma{200}
		\def\ouralpha{2.352941}
		\def\bm3dpsnr{19.13}
		\def\ourspsnr{20.32}
		\begin{subfigure}[t]{\imgwidth}
			\centering
			\includegraphics[width=\textwidth]{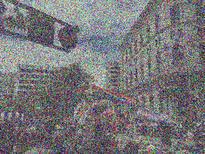}
			\vspace{\capwidth}
			\caption*{\scriptsize input ($\sigma=\imgsigma$)}
		\end{subfigure}	
		\begin{subfigure}[t]{\imgwidth}
			\centering
			\includegraphics[width=\textwidth]{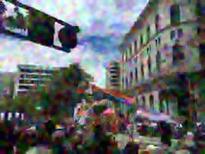}
			\vspace{\capwidth}
			\caption*{\scriptsize BM3D (PSNR=$\bm3dpsnr$)}
		\end{subfigure}
		\begin{subfigure}[t]{\imgwidth}
			\centering
			\includegraphics[width=\textwidth]{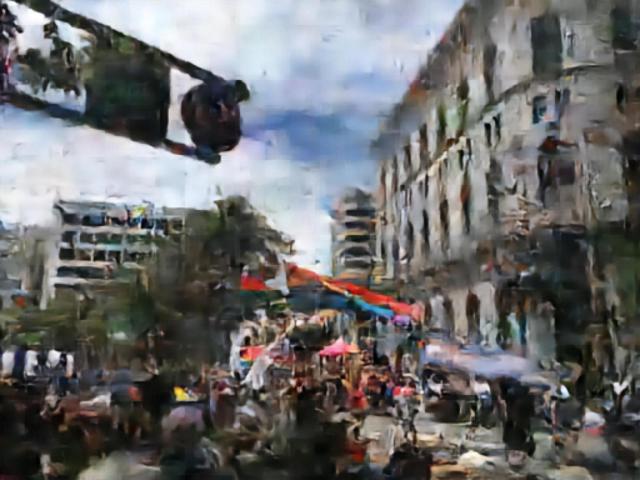}
			\vspace{\capwidth}
			\caption*{\scriptsize proposed (PSNR=$\ourspsnr$)}
		\end{subfigure}
	\end{subfigure}
	\\
	\vspace{3mm}
	\def\idxone{2331}
	\begin{subfigure}{\linewidth}
		\centering
		\begin{subfigure}[t]{\imgwidth}
			\centering
			\includegraphics[width=\textwidth]{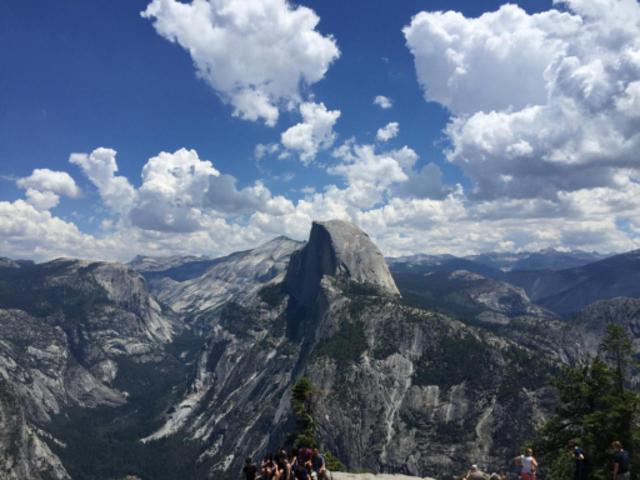}
			\vspace{\capwidth}
			\caption*{\scriptsize ground truth}
		\end{subfigure}	
		\def\imgsigma{40}
		\def\ouralpha{0.470588}
		\def\bm3dpsnr{28.32}
		\def\ourspsnr{29.33}
		\begin{subfigure}[t]{\imgwidth}
			\centering
			\includegraphics[width=\textwidth]{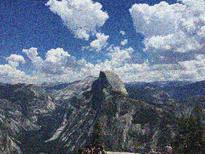}
			\vspace{\capwidth}
			\caption*{\scriptsize input ($\sigma=\imgsigma$)}
		\end{subfigure}	
		\begin{subfigure}[t]{\imgwidth}
			\centering
			\includegraphics[width=\textwidth]{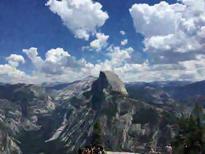}
			\vspace{\capwidth}
			\caption*{\scriptsize BM3D (PSNR=$\bm3dpsnr$)}
		\end{subfigure}
		\begin{subfigure}[t]{\imgwidth}
			\centering
			\includegraphics[width=\textwidth]{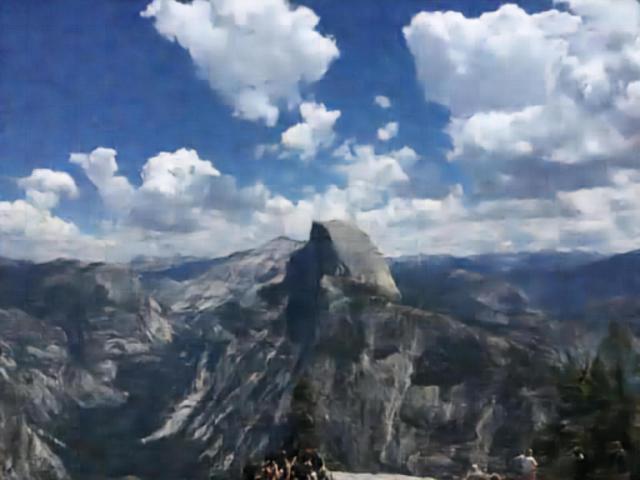}
			\vspace{\capwidth}
			\caption*{\scriptsize proposed (PSNR=$\ourspsnr$)}
		\end{subfigure}
		\\
		\vspace{1mm}
		\begin{subfigure}[t]{\imgwidth}
			\centering
			\includegraphics[width=\textwidth]{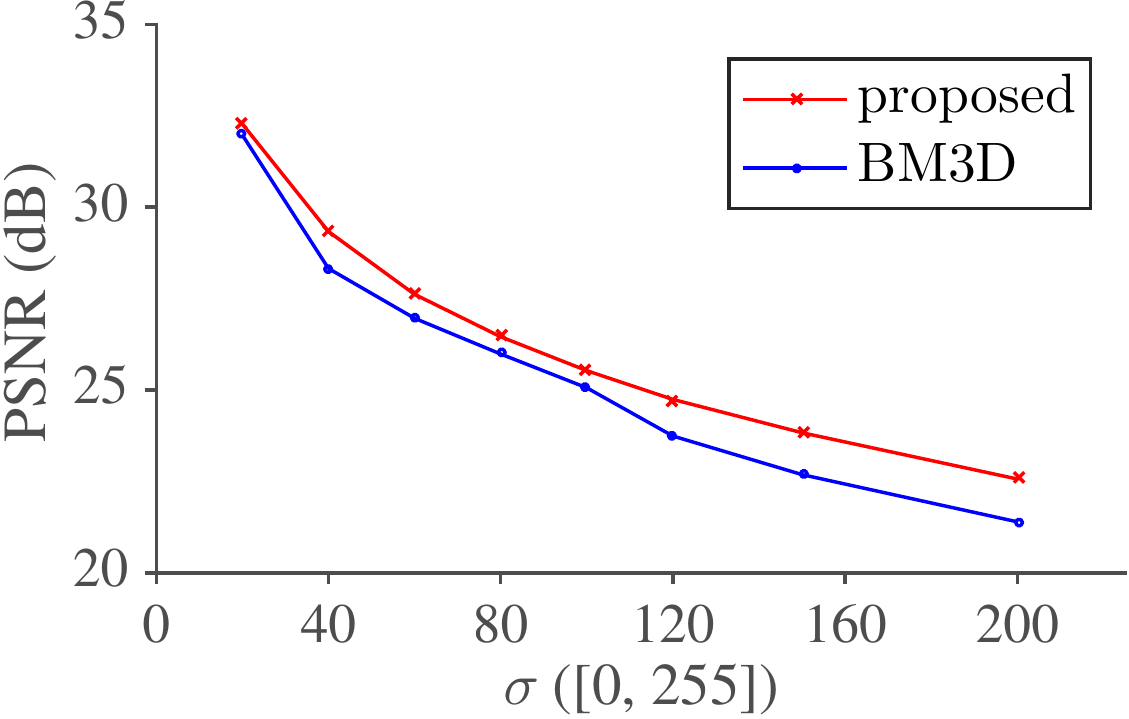}
			\vspace{\capwidth}
			\caption*{\scriptsize PSNR \vs $\sigma$}
		\end{subfigure}	
		\def\imgsigma{100}
		\def\ouralpha{1.176471}
		\def\bm3dpsnr{25.06}
		\def\ourspsnr{25.53}
		\begin{subfigure}[t]{\imgwidth}
			\centering
			\includegraphics[width=\textwidth]{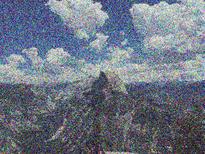}
			\vspace{\capwidth}
			\caption*{\scriptsize input ($\sigma=\imgsigma$)}
		\end{subfigure}	
		\begin{subfigure}[t]{\imgwidth}
			\centering
			\includegraphics[width=\textwidth]{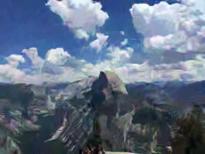}
			\vspace{\capwidth}
			\caption*{\scriptsize BM3D (PSNR=$\bm3dpsnr$)}
		\end{subfigure}
		\begin{subfigure}[t]{\imgwidth}
			\centering
			\includegraphics[width=\textwidth]{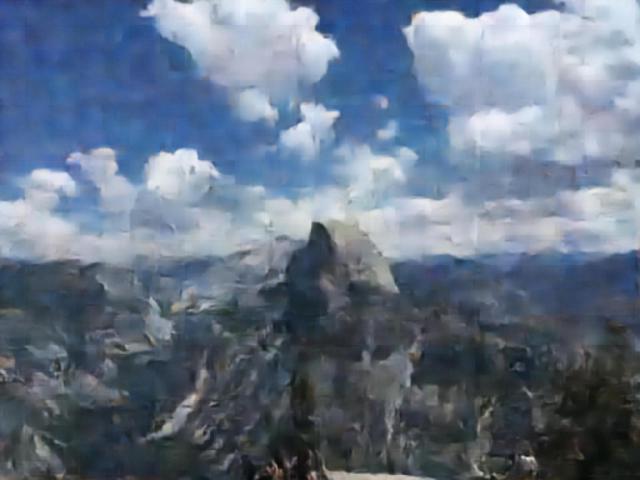}
			\vspace{\capwidth}
			\caption*{\scriptsize proposed (PSNR=$\ourspsnr$)}
		\end{subfigure}
		\\
		\vspace{1mm}
		\begin{subfigure}[t]{\imgwidth}
			\centering
			\hspace{\textwidth}
		\end{subfigure}	
		\def\imgsigma{200}
		\def\ouralpha{2.352941}
		\def\bm3dpsnr{21.40}
		\def\ourspsnr{22.57}
		\begin{subfigure}[t]{\imgwidth}
			\centering
			\includegraphics[width=\textwidth]{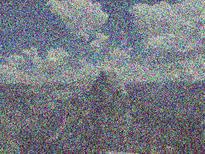}
			\vspace{\capwidth}
			\caption*{\scriptsize input ($\sigma=\imgsigma$)}
		\end{subfigure}	
		\begin{subfigure}[t]{\imgwidth}
			\centering
			\includegraphics[width=\textwidth]{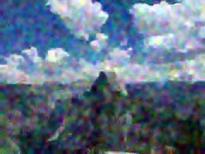}
			\vspace{\capwidth}
			\caption*{\scriptsize BM3D (PSNR=$\bm3dpsnr$)}
		\end{subfigure}
		\begin{subfigure}[t]{\imgwidth}
			\centering
			\includegraphics[width=\textwidth]{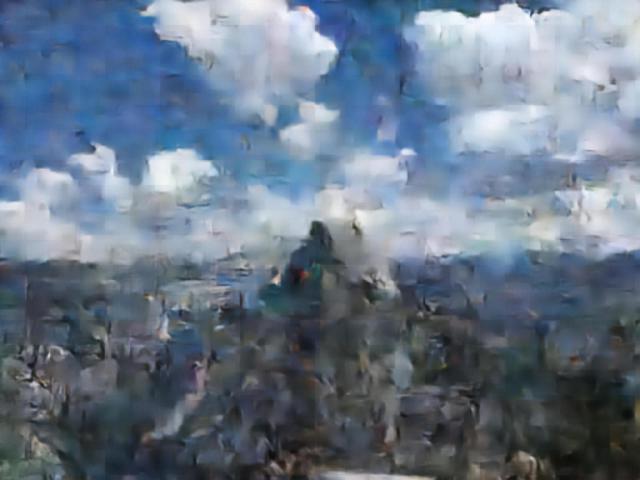}
			\vspace{\capwidth}
			\caption*{\scriptsize proposed (PSNR=$\ourspsnr$)}
		\end{subfigure}
	\end{subfigure}
	\caption{Comparison to BM3D on image denoising}
	\label{figure: denoise}
\end{figure*}

\myparagraph{}
For each of the experiments, we use $\rho=0.3$ if not mentioned.
%
%
The results on MNIST, MS-Celeb-1M, and ImageNet dataset are shown in Figure~\ref{figure: mnist}, Figure~\ref{figure: celeb}, and Figure~\ref{figure: imgnet}, respectively.
In addition, we apply the projection network trained on ImageNet dataset on an image on the Internet~\cite{ring}.  
To deal with the $384 \times 512$ image, when solving the projection operation~\eqref{eq: x update}, we apply the projection network on $64 \times 64$ patches and stitch the results directly. 
The reconstruction outputs are shown in Figure~\ref{figure: venice}, and their statistics of each iteration of ADMM are shown in Figure~\ref{figure: convergence}.

As can be seen from the results, using the proposed projection operator/network learning from datasets enables us to solve more challenging problems than using the traditional wavelet sparsity prior.
In Figure~\ref{figure: mnist} and Figure~\ref{figure: celeb}, while the traditional $\ell_1$-prior of wavelet coefficients is able to reconstruct images from compressive measurements with $\frac{m}{d}=0.3$, it fails to handle larger compression ratios like $\frac{m}{d}=0.1$ and $0.03$.  
Similar observations can be seen on pixelwise inpainting of different dropping probabilities and scattered and blockwise inpainting. 
In contrast, since the proposed projection network is tailored to the datasets, it enables the ADMM algorithm to solve challenging problems like compressive sensing with small $\frac{m}{d}$ and blockwise inpainting on MS-Celeb dataset.

\myparagraph{Robustness to changes in linear operator and to noise.} 
Even though the specially-trained networks are able to generate state-of-the-art results on their designing tasks,  they are unable to deal with similar problems, even with a slight change of the linear operator $A$.  
For example, as shown in Figure~\ref{figure: celeb}, the blockwise inpainting network is able to deal with much larger vacant regions; however, it overfits the problem and fails to fill contents to smaller blocks in scatted inpainting problems.
The $2\times$-super resolution network also fails to reconstruct higher resolution images for $4\times$-super resolution tasks, despite both inputs are upsampled using bicubic algorithm beforehand.
We extend this argument with a compressive sensing example.  
We start from the random Gaussian matrix $A_0$ used to train the compressive sensing network, and we progressively resample elements in $A_0$ from the same distribution constructing $A_0$.
As shown in Figure~\ref{figure: cs compare}, once the portion of resampled elements increases, the specially-trained network fails to reconstruct the inputs, even though the new matrices are still Gaussian.
The network also shows lower tolerant to Gaussian noise added to the clean linear measurements $\y = A_0 \x_0$. 
In comparison, the proposed projector network is robust to changes of linear operators and noise.

\myparagraph{Convergence of ADMM.}
Theorem~\ref{thm: conv} provides a sufficient condition for the nonconvex ADMM to converge.
As discussed in Section~\ref{sec: details}, based on Theorem 1, we use exponential linear units as the activation functions in $\mD$ and $\mD_\ell$ and truncate their weights after each training iteration, in order for the gradient of $\mD$ and $\mD_\ell$ to be Lipschitz continuous.
Even though Theorem~\ref{thm: conv} is just a sufficient condition, in practice, we observe improvement in terms of convergence. 
We conduct experiments on scattered inpainting on ImageNet dataset using two projection networks --- one trained with $\mD$ and $\mD_\ell$ using the smooth exponential linear units, and the other trained with $\mD$ and $\mD_\ell$ using the non-smooth leaky rectified linear units.  
Note that leaky rectified linear units are indifferentiable and thus violate the sufficient condition provided by Theorem~\ref{thm: conv}. 
Figure~\ref{figure: diff vs indiff} shows the root mean square error of $\x - \z$, which is a good indicator of the convergence of ADMM, of the two networks.
As can be seen, using leaky rectified linear units results in higher and spikier root mean square error of $\x - \z$ than using exponential linear units.   
This indicates a less stable ADMM process.  
These results show that following Theorem~\ref{thm: conv} can benefit the convergence of ADMM.

\myparagraph{Failure cases.}
The proposed projection network can fail to solve very challenging problems like the blockwise inpainting on ImageNet dataset, which has higher varieties in image contents than the other two datasets we test on.
As shown in Figure~\ref{figure: imgnet}, the proposed projection network tries to fill in random edges in the missing regions.
In these cases, the projection network fails to project inputs to the natural image set, and thereby, violates our assumption in Theorem~\ref{thm: conv} and affects the overall ADMM framework. 
Even though increasing $\rho$ can improve the convergence, it may produce  low-quality, overly smoothed outputs.

	\section{Conclusion}

In this paper, we propose a general framework to implicitly learn a signal prior --- in the form of a projection operator --- for solving generic linear inverse problems. 
The learned projection operator enjoys the high flexibility of deep neural nets and wide applicability of traditional signal priors.
%
%
With the ability to solve generic linear inverse problems like denoising, inpainting, super-resolution and compressive sensing, the proposed framework resolves the scalability of specially-trained networks.
This characteristic significantly lowers the cost to design specialized hardwares (ASIC for example) to solve image processing tasks.
Thereby, we envision the projection network to be embedded into consumer devices like smart phones and autonomous vehicles  to solve a variety of image processing problems.



	\appendix
	
	\section{Network Architecture}
\label{sec: arch}

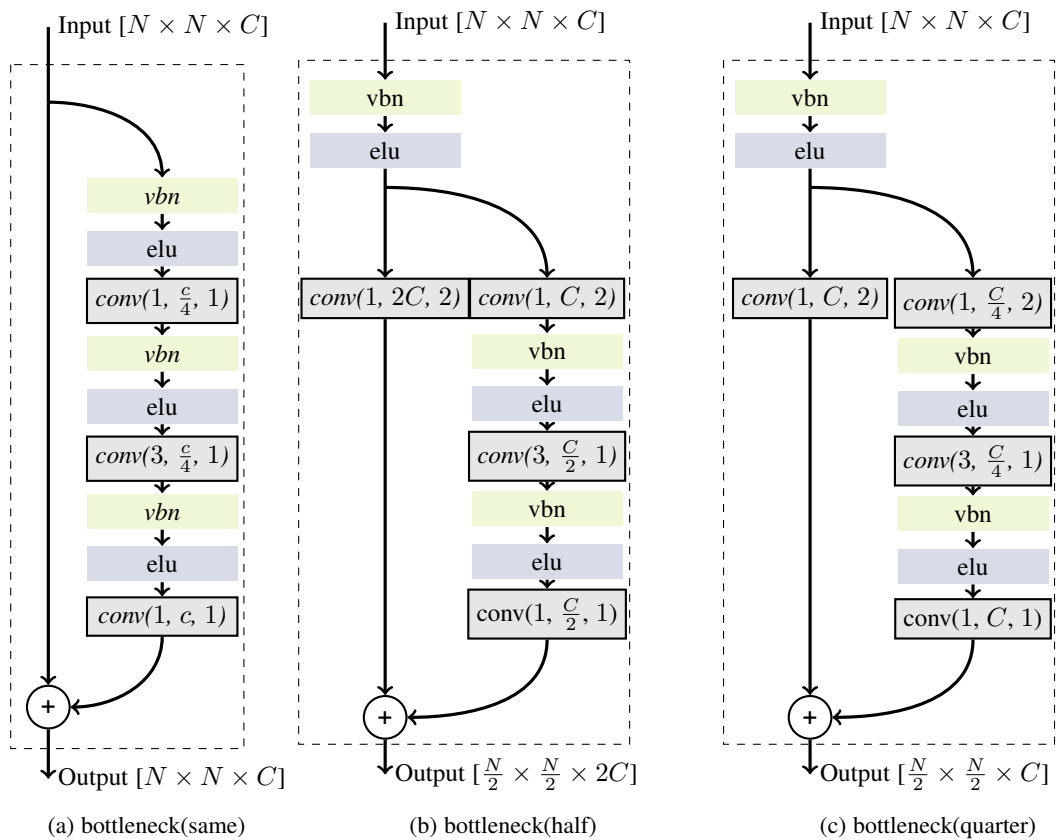
\begin{figure*}
\centering
\definecolor{mycolor1}{HTML}{aadb32}
\definecolor{mycolor2}{HTML}{5bc862}
\definecolor{mycolor3}{HTML}{27ad80}
\definecolor{mycolor4}{HTML}{208f8c}
\definecolor{mycolor5}{HTML}{2c718e}
\definecolor{mycolor6}{HTML}{3b518a}
\definecolor{mycolor7}{HTML}{472b7a}
\begin{subfigure}[t]{0.2\textwidth}
	\begin{tikzpicture}[  
	block/.style    = {draw, thick, rectangle, minimum height = 1.25em, minimum width = 2cm, node distance = 0.7cm, fill=black!10!white},
	relu/.style    = { thick, rectangle, minimum height = 1.25em, minimum width = 2cm, node distance = 0.7cm, fill=mycolor6!20!white},
	batch/.style    = { thick, rectangle, minimum height = 1.25em, minimum width = 2cm, node distance = 0.7cm, fill=mycolor1!20!white},
    sum/.style      = {draw, circle, node distance = 1.cm}, 
	input/.style    = {coordinate, node distance = 1.cm}, 
  	output/.style   = {coordinate, node distance = 1.cm} 
]

\node[input] (input1)  {};
\node[batch,below right=1.cm and 0.5cm of input1] (batch_norm1) {\it vbn};
\node[relu, below of=batch_norm1] (relu1) {elu};
\node[block, below of=relu1]  (layer1)   {\it conv($1$, $\frac{c}{4}$, $1$)};
\node[batch, below of=layer1 ] (batch_norm2) {\it vbn};
\node[relu, below of=batch_norm2] (relu) {elu};
\node[block, below of=relu]  (layer2)   {\it conv($3$, $\frac{c}{4}$, $1$)};
\node[batch, below of=layer2 ] (batch_norm3) {\it vbn};
\node[relu, below of=batch_norm3] (relu3) {elu};
\node[block, below of=relu3]  (layer3)   {\it conv($1$, $c$, $1$)};

\node[sum, below = 7.75cm of input1, thick] (suma) {\textbf{+}};
\node[output, below of=suma] (output1)  {};
 
\draw[->, very thick] (input1) --  (suma);
\draw[->, very thick] (input1) to [out=0, in=90]  (batch_norm1.north);
\draw[->, very thick] (batch_norm1) --  (relu1);
\draw[->, very thick] (relu1) --  (layer1);
\draw[->, very thick] (layer1) --  (batch_norm2);
\draw[->, very thick] (batch_norm2) --  (relu);
\draw[->, very thick] (relu) --  (layer2);
\draw[->, very thick]  (layer2) -- (batch_norm3);
\draw[->, very thick] (batch_norm3) --  (relu3);
\draw[->, very thick] (relu3) --  (layer3);
\draw[->, very thick] (layer3.south) to [out=-90, in=0 ]  (suma.east);

\draw[dashed] ( -0.5,0.5) rectangle (2.6,-8.6);
\draw[-, very thick] (0,1) node[right]{Input [$N\times N\times C$]} -- (input1);
\draw[-> , very thick] (suma) -- (0, -9) node[right]{Output [$N\times N\times C$]};

\end{tikzpicture}
\caption{bottleneck(same)}
\label{fig:resnet_unit}
\end{subfigure}~
\begin{subfigure}[t]{0.3\textwidth}
	\begin{tikzpicture}[  
	block/.style    = {draw, thick, rectangle, minimum height = 1.25em, minimum width = 2cm, node distance = 0.7cm, fill=black!10!white},
	relu/.style    = { thick, rectangle, minimum height = 1.25em, minimum width = 2cm, node distance = 0.7cm, fill=mycolor6!20!white},
	batch/.style    = { thick, rectangle, minimum height = 1.25em, minimum width = 2cm, node distance = 0.7cm, fill=mycolor1!20!white},
    sum/.style      = {draw, circle, node distance = 1.cm}, 
	input/.style    = {coordinate, node distance = 1.cm}, 
  	output/.style   = {coordinate, node distance = 1.cm} 
]

\node[batch] (batch_norm1) {vbn};
\node[relu, below of=batch_norm1] (relu1) {elu};

\node[block, below right=1.45cm and 0.1cm of relu1]  (layer1)   {\it conv($1$, $C$, $2$)};
\node[block, below = 1.45cm of relu1]  (proj1)   {\it conv($1$, $2C$, $2$)};
\node[batch, below of=layer1 ] (batch_norm2) {vbn};
\node[relu, below of=batch_norm2] (relu) {elu};
\node[block, below of=relu]  (layer2)   {\it conv($3$, $\frac{C}{2}$, $1$)};
\node[batch, below of=layer2 ] (batch_norm3) {vbn};
\node[relu, below of=batch_norm3] (relu3) {elu};
\node[block, below of=relu3]  (layer3)   {conv($1$, $\frac{C}{2}$, $1$)};
\node[sum, below = 5.cm of proj1, thick] (suma) {\textbf{+}};
\node[output, below of=suma] (output1)  {};
 
\draw[->, very thick] (proj1) --  (suma);
\draw[->, very thick] (relu1.south) ++ (0,-0.25) to [out=0, in=90]  (layer1.north);
\draw[->, very thick] (batch_norm1) --  (relu1);
\draw[->, very thick] (relu1) --  (proj1);
\draw[->, very thick] (layer1) --  (batch_norm2);
\draw[->, very thick] (batch_norm2) --  (relu);
\draw[->, very thick] (relu) --  (layer2);
\draw[->, very thick]  (layer2) -- (batch_norm3);
\draw[->, very thick] (batch_norm3) --  (relu3);
\draw[->, very thick] (relu3) --  (layer3);
\draw[->, very thick] (layer3.south) to [out=-90, in=0 ]  (suma.east);

\draw[dashed] ( -1.15,0.5) rectangle (3.25,-8.6);
\draw[->, very thick] (0,1) node[right]{Input [$N\times N\times C$]} -- (batch_norm1);
\draw[->, very thick] (suma) -- (0, -9) node[right]{Output [$\frac{N}{2}\times \frac{N}{2} \times 2C$]};
\end{tikzpicture}
\caption{bottleneck(half)}
\label{fig:resnet_unit_dec}
\end{subfigure}~
\begin{subfigure}[t]{0.3\textwidth}
	\begin{tikzpicture}[  
	block/.style    = {draw, thick, rectangle, minimum height = 1.25em, minimum width = 2cm, node distance = 0.7cm, fill=black!10!white},
	relu/.style    = { thick, rectangle, minimum height = 1.25em, minimum width = 2cm, node distance = 0.7cm, fill=mycolor6!20!white},
	batch/.style    = { thick, rectangle, minimum height = 1.25em, minimum width = 2cm, node distance = 0.7cm, fill=mycolor1!20!white},
    sum/.style      = {draw, circle, node distance = 1.cm}, 
	input/.style    = {coordinate, node distance = 1.cm}, 
  	output/.style   = {coordinate, node distance = 1.cm} 
]

\node[batch] (batch_norm1) {vbn};
\node[relu, below of=batch_norm1] (relu1) {elu};

\node[block, below right=1.45cm and 0.1cm of relu1]  (layer1)   {\it conv($1$, $\frac{C}{4}$, $2$)};
\node[block, below = 1.45cm of relu1]  (proj1)   {\it conv($1$, $C$, $2$)};
\node[batch, below of=layer1 ] (batch_norm2) {vbn};
\node[relu, below of=batch_norm2] (relu) {elu};
\node[block, below of=relu]  (layer2)   {\it conv($3$, $\frac{C}{4}$, $1$)};
\node[batch, below of=layer2 ] (batch_norm3) {vbn};
\node[relu, below of=batch_norm3] (relu3) {elu};
\node[block, below of=relu3]  (layer3)   {conv($1$, $C$, $1$)};
\node[sum, below = 5.cm of proj1, thick] (suma) {\textbf{+}};
\node[output, below of=suma] (output1)  {};
 
\draw[->, very thick] (proj1) --  (suma);
\draw[->, very thick] (relu1.south) ++ (0,-0.25) to [out=0, in=90]  (layer1.north);
\draw[->, very thick] (batch_norm1) --  (relu1);
\draw[->, very thick] (relu1) --  (proj1);
\draw[->, very thick] (layer1) --  (batch_norm2);
\draw[->, very thick] (batch_norm2) --  (relu);
\draw[->, very thick] (relu) --  (layer2);
\draw[->, very thick]  (layer2) -- (batch_norm3);
\draw[->, very thick] (batch_norm3) --  (relu3);
\draw[->, very thick] (relu3) --  (layer3);
\draw[->, very thick] (layer3.south) to [out=-90, in=0 ]  (suma.east);

\draw[dashed] ( -1.15,0.5) rectangle (3.25,-8.6);
\draw[->, very thick] (0,1) node[right]{Input [$N\times N\times C$]} -- (batch_norm1);
\draw[->, very thick] (suma) -- (0, -9) node[right]{Output [$\frac{N}{2}\times \frac{N}{2} \times C$]};
\end{tikzpicture}
\caption{bottleneck(quarter)}
\label{fig:bottle_double}
\end{subfigure}

	\caption{Block diagrams of the bottleneck components used in the paper. (a) \textit{bottleneck(same)} preserves the dimensionality of the input by maintaining the same output spatial dimension and the numger of channels. (b) \textit{bottleneck(half)} reduces dimensionality by $2$ via halving each spatial dimension and doubling the number of channels.  (c) \textit{bottleneck(quarter)} reduces dimensionality by $4$ via halving each spatial dimension. }
	\label{fig:residual_block}
\end{figure*}

We now describe the architecture of the networks used in the paper.
We use exponential linear unit (elu)~\cite{clevert2015fast} as activation function. 
We also use virtual batch normalization~\cite{salimans2016improved}, where the reference batch size $b_{\textrm{ref}}$ is
equal to the batch size used for stochastic gradient descent.   
We weight the reference batch with $\frac{b_{\textrm{ref}}}{b_{\textrm{ref}} + 1}$.
We define some shorthands for the basic components used in the networks.
\begin{itemize}
	\item {\it conv(w, c, s)}: convolution with $w\times w$ window size, $c$ output channels and $s$ stride. 
	\item {\it dconv(w, c, s)} deconvolution (transpose of the convolution operation) with $w\times w$ window size, $c$ output channels and $s$ stride. 
	\item {\it vbn}: virtual batch normalization.
	\item {\it bottleneck(same/half/quarter)}: bottleneck residual units~\cite{he2016res} having the same, half, or one-fourth of the dimensionality of the input.  Their block diagrams are shown in Figure~\ref{fig:residual_block}.  
	\item {\it cfc}: a channel-wise fully connected layer, whose output dimension is with same size as the input
	dimension.
	\item {\it fc(s)}: a fully-connected layer with the output size $s$.
\end{itemize}
To simply the notation, we use the subscript {\it ve} on a component to indicate that it is followed by {\it vbn} and {\it elu}. 

\paragraph{Projection network $\mP$.}

The projection network $\mP$ is composed of one encoder network $\mE$ and one decoder network, like a typical autoencoder. 
The encoder $\mE$ projects an input to a $1024$-dimensional latent space, and the decoder projects the latent representation back to the image space.
The architecture of $\mE$ is as follows. 
\begin{equation}
\begin{array}{ccccc}
 Input & \rightarrow &  conv(4,64,1)_{ve} & \rightarrow & conv(4,128,1)_{ve}  \\
 & \rightarrow & conv(4,256,2)_{ve} & \rightarrow & conv(4,512, 2)_{ve} \\
 & \rightarrow &  conv(4,1024,2)_{ve} &\rightarrow & \mathit{cfc}  \\
 & \rightarrow &  conv(2,1024,1)_{ve} & & (latent) 
\end{array}
\label{eq:encoder}
\end{equation}

The decoder is a symmetric counter part of the encoder:
\begin{equation}
\begin{array}{ccccc}
	latent & \rightarrow & dconv(4,512,2)_{ve} & \rightarrow & dconv(4,256,2)_{ve}  \\
	& \rightarrow & dconv(4,128,1)_{ve} & \rightarrow & dconv(4,64,1)_{ve}  \\
	& \rightarrow & dconv(4,3,1)_& & (Output) 
\end{array}
\end{equation}

\paragraph{Image-space classifier $\mD$.}
	
As shown in Figure~$3$ of the paper, we use two classifiers --- one operates in the image space $\R^d$ and discriminates natural images from the projection outputs, the other operates in the latent space of $\mP$ based on the hypothesis that after encoded by $\mE$, a perturbed image and a natural image should already lie in the same set.   

For the image-space classifier $\mD$, we use the $50$-layer architecture of~\cite{he2016deep} but use the bottleneck blocks suggested in~\cite{he2016res}.
The detailed architecture is as follows.
\begin{equation}
\begin{array}{ccccc}
	Input & \rightarrow & conv(4,64,1) & &  \\
	& \rightarrow & \mathit{bottleneck(half)} & \rightarrow & \{\mathit{bottleneck(same)}\}_{\times 3}  \\
	& \rightarrow & \mathit{bottleneck(half)} & \rightarrow & \{\mathit{bottleneck(same)}\}_{\times 4}  \\
	& \rightarrow & \mathit{bottleneck(half)} & \rightarrow & \{\mathit{bottleneck(same)}\}_{\times 6}  \\
	& \rightarrow & \mathit{bottleneck(half)} & \rightarrow & \{\mathit{bottleneck(same)}\}_{\times 3}  \\
	& \rightarrow & vbn\mbox{ }\&~elu & \rightarrow & \mathit{fc}(1)~(output),
\end{array}
\end{equation}
where $\{\}_{\times n}$ means we repeat the building block $n$ times.

\paragraph{Latent-space classifier $\mD_\ell$.}
The latent space classifier~$\mD_\ell$ operates on the output of the encoder~$\mE$. 
Since the input dimension is smaller than that of $\mD$, we use fewer $bottleneck$ blocks than we did in $\mD$.

\begin{equation}
\begin{array}{ccc}
	Input & \rightarrow  & \mathit{bottleneck(same)}_{\times 3}   \\
	& \rightarrow & \mathit{bottleneck(quarter)} \\
	& \rightarrow & \{\mathit{bottleneck(same)}\}_{\times 2} \\
	& \rightarrow & vbn~\&~elu \\ 
	& \rightarrow & \mathit{fc}(1)~(output)

\end{array}
\end{equation}


		\FloatBarrier


	\bibliographystyle{IEEEtran}
	\bibliography{egbib}

\end{document}